\documentclass[twocolumn, 10pt]{article}

\pdfoutput=1

\usepackage{abstract}
\usepackage{amsmath}
\usepackage{amsthm}
\usepackage{bbm}
\usepackage{bibunits}
\usepackage{booktabs}
\usepackage[inline]{enumitem}
\usepackage{etoolbox}
\usepackage[font=small,labelfont=sc]{caption}
\usepackage[T1]{fontenc}
\usepackage{fullpage}
\usepackage{graphicx}
\usepackage[colorlinks=true,
            linkcolor=blue,
            citecolor=blue,
            urlcolor=black]{hyperref}
\usepackage{lmodern}
\usepackage{makecell}
\usepackage{multirow}
\usepackage[square]{natbib}
\usepackage{placeins}
\usepackage[detect-all]{siunitx}
\usepackage{subcaption}
\usepackage{tabularx}
\usepackage[table]{xcolor}
\usepackage{titling}
\usepackage{xr}

\defaultbibliography{leaps}
\defaultbibliographystyle{plainnat}

\robustify\bfseries

\definecolor{lightgray}{gray}{0.9}

\let\oldtabular\tabular
\renewcommand{\tabular}{\small\oldtabular}

\title{Modeling Industrial ADMET Data with Multitask Networks}

\author{
    \textbf{\normalsize Steven Kearnes} \\
    \normalsize Stanford University \\
    \texttt{\normalsize kearnes@stanford.edu} \and
    \textbf{\normalsize Brian Goldman} \\
    \normalsize Vertex Pharmaceuticals Inc. \\
    \texttt{\normalsize brian\_goldman@vrtx.com} \and
    \textbf{\normalsize Vijay Pande} \\
    \normalsize Stanford University \\
    \texttt{\normalsize pande@stanford.edu} \and
}

\date{}  

\AtBeginDocument{}

\begin{document}

\maketitle
\begin{bibunit}
\begin{abstract}

Deep learning methods such as multitask neural networks have recently been
applied to ligand-based virtual screening and other drug discovery applications.
Using a set of industrial ADMET datasets, we compare neural networks to standard
baseline models and analyze multitask learning effects with both random
cross-validation and a more relevant temporal validation scheme. We confirm that
multitask learning can provide
modest benefits over single-task models and show that smaller datasets tend to
benefit more than larger datasets from multitask learning. Additionally, we
find that adding massive amounts of side information is not guaranteed to
improve performance relative to simpler multitask learning.
Our results emphasize that multitask effects are highly dataset-dependent,
suggesting the use of dataset-specific models to maximize overall performance.

\end{abstract}


\section{Introduction}
\label{sec:leaps_intro}

The 2012 Merck Molecular Activity Challenge~\citep{dahl2012deep} catalyzed a
surge of interest in using artificial neural networks and ``deep
learning''~\citep{lecun2015deep} for problems in drug discovery and
cheminformatics, especially virtual screening. Follow-up studies have shown that
neural networks, on average, outperform traditional machine learning methods
such as random forest~\citep{dahl2014multi, ma2015deep, ramsundar2015massively,
mayr2015deeptox}.

\begin{figure}[tb]
  \centering
  \includegraphics[clip,trim=0 245 350 0,width=\linewidth]{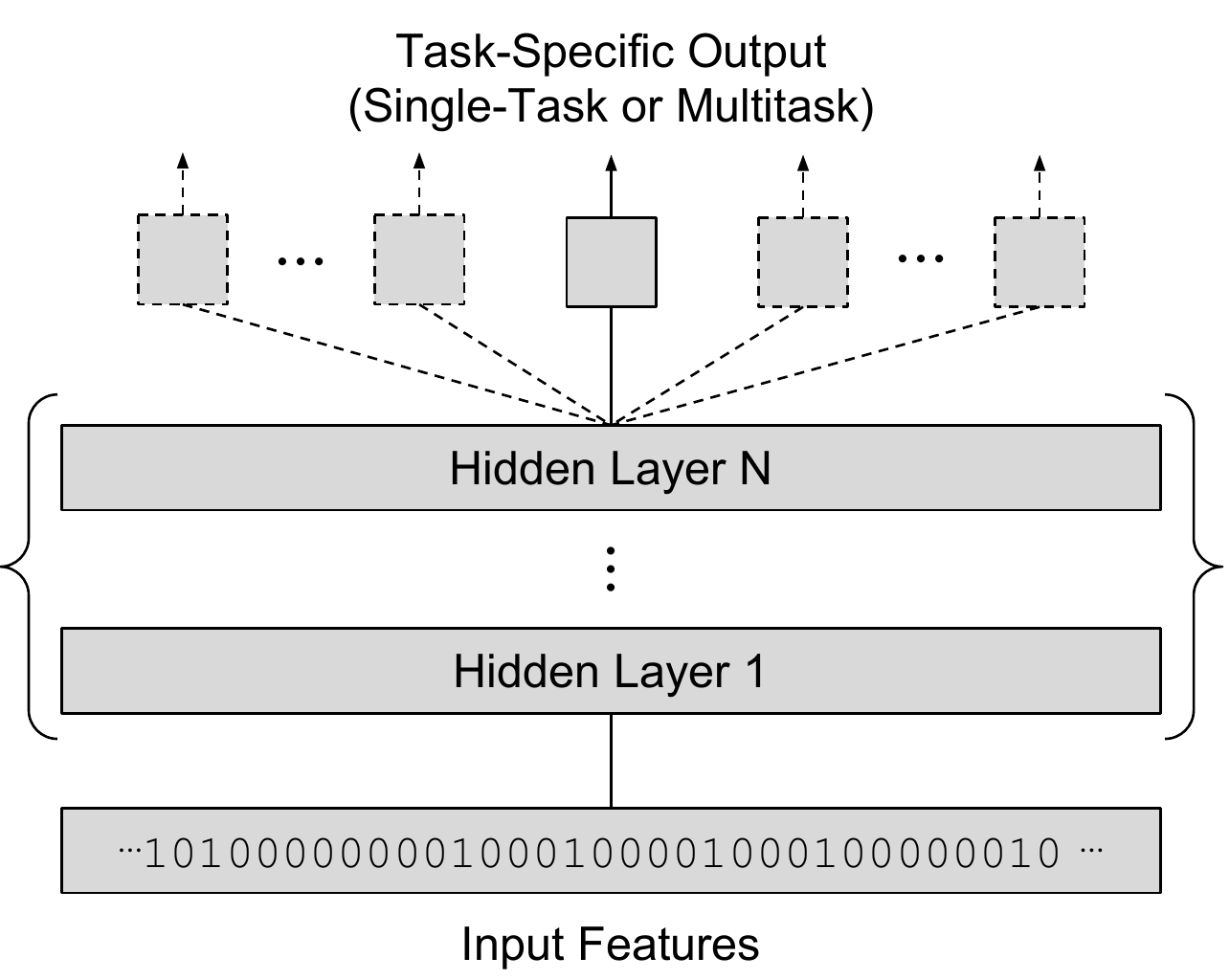}
  \caption{Abstract neural network architecture. The input vector is a binary
  molecular fingerprint with 1024 bits. All connections between layers are \emph{dense},
  meaning that every unit in layer $n$ is connected to every unit in layer
  ${n+1}$. Each output block is a task-specific two-class softmax layer; dashed
  lines indicate that models can be either single-task or multitask.}
  \label{fig:network_diagram}
\end{figure}

The winners of the Merck challenge utilized multitask neural networks
(MTNNs), which are trained to predict many outputs simultaneously in order to
improve performance relative to single-task models~\citep{caruana1997multitask}.
As shown in \figurename~\ref{fig:network_diagram}, MTNNs share a subset of model
parameters between all tasks, encouraging the model to learn an internal
representation of the input data that is useful for all tasks. If the tasks in a
multitask model are related to one another, the parameters in the hidden
representation are effectively exposed to more training data than they would be
in a single-task model with the same architecture. This ``data amplification''
effect---possibly in combination with additional effects described
by~\citet{caruana1997multitask}---helps to explain how multitask models can
outperform their single-task analogs (the so-called \emph{multitask effect}).

Although multitask learning improves performance on average, it is not clear
what factors contribute to multitask improvement for specific datasets. One
major issue is that task \emph{relatedness} is not well defined.
For example, \citet{ma2015deep} observed that models trained on the largest
datasets in their collection, which had many molecules in common, performed
\emph{worse} in a multitask setting. Additionally, the amount of data required
for multitask learning is not well understood, nor is it possible to predict
when training with additional data or tasks will significantly improve
performance. Some
studies using public data have emphasized the addition of large amounts of side
information to improve multitask performance~\citep{ramsundar2015massively,
unterthiner2014deep}, leading to the impression that more data is better (on
average) with little dependence on apparent task relatedness or dataset size.

Apart from uncertainty surrounding the multitask effect, it is possible that
results based on publicly available data will not transfer well to
industrial settings. Besides data quality, the most obvious potential issue
is the use of random cross-validation instead of temporal validation.
\citet{sheridan2013time} demonstrated that random cross-validation generally
gives overly optimistic estimates of prospective performance, but temporal
validation is often impossible when working with public data due to the lack of
temporal metadata (timestamps). In this context, \citet{ma2015deep} used temporally annotated
industrial data
to demonstrate that neural networks can outperform random forest models (they
also performed some experiments with multitask networks).

This report describes applications of single-task and multitask neural networks
to absorption, distribution, metabolism, excretion, and/or toxicity (ADMET) assay data in an
industrial drug discovery setting at Vertex Pharmaceuticals. Using a selection
of our internal datasets, our main objectives
are to (1) compare single-task and
multitask neural network models to baseline models including random forest and
logistic regression, (2) assess the
impact of temporal validation \emph{vs.} random cross-validation on model
performance, and (3) explore several factors that could potentially influence multitask effects.
Additionally, we consider the problem of information leakage in multitask
models.
Our results confirm previous work showing that multitask learning can improve model performance while
emphasizing the fact that multitask effects are highly dataset-dependent.

\section{Methods}

\subsection{Datasets}
\label{sec:datasets}

Our experiments focused on classification performance for a set of 22 Vertex
datasets with various ADMET endpoints including hERG inhibition, aqueous
solubility, compound metabolism, and others. Experimental data points were
divided into ``active'', ``inconclusive'', and ``inactive'' classes using
dataset-specific thresholds, and ``inconclusive'' results were discarded. These
datasets contained approximately \num{280000} experimental values and are
described in \tablename~\ref{table:datasets}. Additional data used as side
information (see Section~\ref{sec:total_data}) are described in
Section~\ref{sec:appendix:datasets}.

Since many of the tasks have imbalanced active/inactive proportions, we weighted
the examples in the minority class such that the active and inactive classes had
equal weight. Active/inactive ratios were calculated using training data (except
for random cross-validation, where the ratio was determined with the full
dataset). Note that these example weights were not used in logistic regression
or random forest models.

\begin{table}[tb]
    \caption{Proprietary datasets used for model evaluation. Each data point is
    associated with an experiment date used for temporal validation.}
    \label{table:datasets}
    \centering
    \rowcolors{1}{}{lightgray}
    \sisetup{table-format=9}
    \begin{tabular}{ c S S S }
    \toprule
    Dataset &
    {Actives} &
    {Inactives} &
    {Total} \\
    \midrule
    A & 20247 & 9652 & 29899 \\
    B & 32806 & 23936 & 56742 \\
    C & 40136 & 27703 & 67839 \\
    D & 24379 & 2374 & 26753 \\
    E & 21722 & 2746 & 24468 \\
    F & 25202 & 2034 & 27236 \\
    G & 2003 & 3226 & 5229 \\
    H & 500 & 526 & 1026 \\
    I & 669 & 344 & 1013 \\
    J & 883 & 399 & 1282 \\
    K & 845 & 357 & 1202 \\
    L & 489 & 164 & 653 \\
    M & 820 & 357 & 1177 \\
    N & 1420 & 740 & 2160 \\
    O & 670 & 1417 & 2087 \\
    P & 3861 & 4107 & 7968 \\
    Q & 1056 & 2658 & 3714 \\
    R & 215 & 2760 & 2975 \\
    S & 987 & 582 & 1569 \\
    T & 1454 & 5935 & 7389 \\
    U & 3998 & 2790 & 6788 \\
    V & 2795 & 896 & 3691 \\
    \midrule
    \rowcolor{white}  
      & 187157 & 95703 & 282860 \\
    \bottomrule
    \end{tabular}
\end{table}

Models were trained using a temporal validation scheme: each data point has an
associated experiment date, and divisions into training, validation, and test
set divisions were based on temporally ordered date ranges as
in~\citet{sheridan2013time}. Date cutoffs were chosen to divide the data for
each dataset into approximately 70\% training, 10\% validation, and 20\% test.
For comparison to temporal models, we also trained models using a random
cross-validation scheme where each dataset was randomly partitioned into five
folds in a stratified manner such that the active/inactive ratio of the full
dataset was approximately preserved in each fold.

Dataset compounds were represented by 1024-bit circular fingerprints with radius
2 (similar to the ECFP4 fingerprints described by~\citet{rogers2010extended})
generated with the~\citet{oegraphsimtk}.

\subsection{Model training and evaluation}

Neural network models were constructed with TensorFlow~\citep{abaditensorflow},
an open source library for machine learning. We used standard feed-forward
architectures with rectified linear activations, batch
normalization~\citep{ioffe2015batch}, 0.5 dropout~\citep{srivastava2014dropout},
and the adagrad optimizer~\citep{duchi2011adaptive} with learning rate 0.001 and
batch size 128. Models were trained for 1~M steps (except as otherwise noted)
with periodic checkpointing. Input examples were encoded in a ``dense'' format,
such that there was one training example for each unique molecule identifier
containing features for that molecule and labels and weights for each task. For
tasks where an example did not have activity data, the weight was set to zero.
Note that we did not use SMILES strings as molecule identifiers, so some
molecules with identical SMILES may have been represented by multiple training
examples.
Models were trained on a cluster with several NVIDIA Tesla K80 GPUs, although no
multi-GPU training was performed.

Each neural network model was based on one of five chosen architectures.
Each architecture is specified using $(x_1, x_2, \ldots, x_n)$ notation, where
the input is fed into a fully-connected layer with $x_1$ units, followed by a
layer with $x_2$ units, and so on. The
(1000) architecture represents a reasonable baseline network. The (2000,~100)
and (2000,~1000) architectures examine how the size of the final representation
affects learning; \citet{ramsundar2015massively} hypothesized that a pyramidal
architecture with relatively few task-specific parameters helps to avoid
overfitting while still providing a rich shared representation. The
(4000,~2000,~1000,~1000) architecture was recommended by \citet{ma2015deep} for
regression on chemical datasets, and the (4000) architecture investigates the
use of a single wide layer in lieu of multiple hidden layers.

Logistic regression and random forest baselines were built using the
\texttt{LogisticRegression} and \texttt{RandomForestClassifier} classes,
respectively, in scikit-learn~\citep{pedregosa2011scikit}. Parameters for random
forest models were similar to those described by~\citet{ma2015deep};
specifically \texttt{n\_estimators=100}, \texttt{max\_features=1/3.}, and
\texttt{min\_samples\_split=6}. We note that these parameters were selected for
different molecular descriptors and may not be optimal for use with circular
fingerprints.

Model performance was evaluated using the area under the receiver operating
characteristic curve (ROC AUC, or simply AUC), which is a global measure of
classification performance~\citep{jain2008recommendations}. For evaluation of
neural network models, a single training checkpoint was selected that maximized
the validation set AUC for the task of interest, and this checkpoint was
used to make predictions and calculate AUC scores for test set compounds. In
practice, this meant that multitask models were often evaluated at several
different training checkpoints that optimized validation set performance for
individual tasks. We note that some tasks achieved their best validation score
near the beginning or end of training and therefore some reported results may be
subject to overfitting or underfitting, respectively.

Although ROC AUC is the recommended metric for evaluating virtual screening
models~\citep{jain2008recommendations}, it is a global metric that is not always
useful for making decisions in a production setting. AUC values reflect the
relative positions of active and inactive compounds in a ranked list of
predicted values, but compound-specific decisions require the choice of a
threshold on predicted values and estimates of prediction confidence.
Additionally, ``enrichment'' scores based on ROC curves can be used to measure
performance early in a ranked list~\citep{jain2008recommendations}.
The analysis that follows is based on ROC AUC scores; explorations of
alternative metrics or strategies for choosing decision thresholds, estimating
prediction confidence, and/or determining the domain of applicability for models
are beyond the scope of this report.

\subsection{Model comparison}
\label{sec:model_comparison}

For comparisons between models we report the median $\Delta$AUC across the
datasets in \tablename~\ref{table:datasets} and a 95\% confidence interval for
the sign test statistic. The sign test is a paired non-parametric test that
measures the fraction of per-dataset $\Delta$AUC values that are greater than
zero (exactly zero differences are excluded). The 95\% confidence interval is a
Wilson score interval around this fraction that estimates the probability that
one model will outperform another (in terms of AUC).
Confidence intervals that do not include 0.5 indicate
statistically significant differences; i.e.~two models are statistically
indistinguishable if there is not a clear bias toward one model or the other.
Conceptually, the sign test confidence interval estimates the consistency of
observed differences between models, while the median $\Delta$AUC estimates the
effect size.
Confidence intervals were calculated with the \texttt{proportion\_confint}
method in statsmodels~\citep{seabold2010statsmodels} using \texttt{alpha=0.05}
and \texttt{method=`wilson'}.

\section{Results}

\subsection{Model performance}
\label{sec:model_performance}

We trained single-task neural network (STNN) and multitask neural network (MTNN)
models on our dataset collection using a temporal validation scheme. We used two
flavors of MTNN: standard MTNNs used uniform weights for the cost associated
with each task (U-MTNN), while task-weighted MTNNs assigned weights to each task
that were inversely proportional to the number of training compounds for each
task, such that the total weight for each task was approximately equal (W-MTNN). This
approach allowed us to investigate whether it is important to upweight small
tasks in order for them to benefit from multitask learning. Additionally, NN
models (but not logistic regression or random forest models) used per-example
weights that attempted to compensate for imbalance between actives and inactives
in each dataset (see Section~\ref{sec:datasets}).

Median test set AUC values for these models are reported in
\tablename~\ref{table:main_results}, along with values for random forest and logistic
regression baselines. For comparisons between models, we report median
$\Delta$AUC values and sign test 95\% confidence intervals (see
Section~\ref{sec:model_comparison}). MTNN models had consistently
better performance than random forest regardless of architecture (note that
``consistent'' does not necessarily imply superior performance for all
datasets). Surprisingly, neural networks were not as robust
when compared to logistic regression; the (2000,~100) and
(4000,~2000,~1000,~1000) MTNNs were statistically indistinguishable
from the logistic regression baseline. However, the (1000), (4000), and
(2000,~1000) MTNN models outperformed logistic regression, with
task-weighted models giving the most consistent improvements
as measured by sign test confidence intervals.

\begin{table*}[tbp]
    \caption{Median test set AUC values for random forest, logistic regression,
    single-task neural network (STNN), and multitask neural network (MTNN)
    models. U-MTNN models treat each task uniformly; W-MTNN models are task-weighted models, meaning that the cost for
    each task is weighted inversely proportional to the amount of training data
    for that task. We also report median $\Delta$AUC values and sign test 95\%
    confidence intervals for comparisons between each model and random forest or
    logistic regression (see Section~\ref{sec:model_comparison}). Bold values
    indicate confidence intervals that do not include 0.5.}
    \label{table:main_results}
    \centering
    \rowcolors{2}{lightgray}{}
    \sisetup{detect-weight=true,detect-inline-weight=math}
    \begin{tabular}{ l l S S c S c }
    \toprule
     & & & \multicolumn{2}{c}{Model - Random Forest} &
           \multicolumn{2}{c}{Model - Logistic Regression} \\
    \cmidrule(lr){4-5} \cmidrule(lr){6-7}
     & Model &
    {\makecell{Median \\ AUC}} &
    {\makecell{Median \\ $\Delta$AUC}} & \makecell{Sign Test \\ 95\% CI} &
    {\makecell{Median \\ $\Delta$AUC}} & \makecell{Sign Test \\ 95\% CI} \\
    \midrule
    \cellcolor{white} & Random Forest & 0.719 &  &  & -0.016 & (\num{0.20}, \num{0.57}) \\
    \cellcolor{white} & Logistic Regression & 0.758 & 0.016 & (\num{0.43}, \num{0.80}) &  &  \\
    \midrule
    \cellcolor{white} & (1000) & 0.748 & 0.043 & (\num{0.47}, \num{0.84}) & 0.007 & (\num{0.39}, \num{0.77}) \\
    \cellcolor{white} & (4000) & 0.761 & \bfseries 0.052 & \bfseries (\num{0.52}, \num{0.87}) & \bfseries 0.015 & \bfseries (\num{0.52}, \num{0.87}) \\
    \cellcolor{white} & (2000, 100) & 0.749 & 0.039 & (\num{0.47}, \num{0.84}) & 0.007 & (\num{0.35}, \num{0.73}) \\
    \cellcolor{white} & (2000, 1000) & 0.759 & 0.038 & (\num{0.47}, \num{0.84}) & 0.008 & (\num{0.35}, \num{0.73}) \\
    \multirow{-5}{*}{\cellcolor{white} STNN} & (4000, 2000, 1000, 1000) & 0.736 & 0.041 & (\num{0.43}, \num{0.80}) & -0.011 & (\num{0.27}, \num{0.65}) \\
    \midrule
    \cellcolor{white} & (1000) & 0.792 & \bfseries 0.049 & \bfseries (\num{0.67}, \num{0.95}) & \bfseries 0.029 & \bfseries (\num{0.52}, \num{0.87}) \\
    \cellcolor{white} & (4000) & 0.768 & \bfseries 0.057 & \bfseries (\num{0.61}, \num{0.93}) & \bfseries 0.031 & \bfseries (\num{0.57}, \num{0.90}) \\
    \cellcolor{white} & (2000, 100) & 0.797 & \bfseries 0.044 & \bfseries (\num{0.61}, \num{0.93}) & 0.023 & (\num{0.43}, \num{0.80}) \\
    \cellcolor{white} & (2000, 1000) & 0.800 & \bfseries 0.071 & \bfseries (\num{0.67}, \num{0.95}) & \bfseries 0.040 & \bfseries (\num{0.52}, \num{0.87}) \\
    \multirow{-5}{*}{\cellcolor{white} U-MTNN} & (4000, 2000, 1000, 1000) & 0.809 & \bfseries 0.059 & \bfseries (\num{0.72}, \num{0.97}) & 0.024 & (\num{0.43}, \num{0.80}) \\
    \midrule
    \cellcolor{white} & (1000) & 0.793 & \bfseries 0.059 & \bfseries (\num{0.78}, \num{0.99}) & \bfseries 0.040 & \bfseries (\num{0.67}, \num{0.95}) \\
    \cellcolor{white} & (4000) & 0.773 & \bfseries 0.055 & \bfseries (\num{0.72}, \num{0.97}) & \bfseries 0.036 & \bfseries (\num{0.67}, \num{0.95}) \\
    \cellcolor{white} & (2000, 100) & 0.769 & \bfseries 0.050 & \bfseries (\num{0.61}, \num{0.93}) & 0.022 & (\num{0.43}, \num{0.80}) \\
    \cellcolor{white} & (2000, 1000) & 0.821 & \bfseries 0.077 & \bfseries (\num{0.78}, \num{0.99}) & \bfseries 0.041 & \bfseries (\num{0.67}, \num{0.95}) \\
    \multirow{-5}{*}{\cellcolor{white} W-MTNN} & (4000, 2000, 1000, 1000) & 0.800 & \bfseries 0.071 & \bfseries (\num{0.61}, \num{0.93}) & 0.035 & (\num{0.47}, \num{0.84}) \\
    \bottomrule
    \end{tabular}
\end{table*}

\tablename~\ref{table:mtnn_vs_stnn} compares single-task and multitask neural
network models with the same core (hidden layer) architecture. Of the MTNN
models we trained, only the (1000) and (2000,~1000) W-MTNN models showed
significant improvement over their single-task counterparts. There were no
consistent differences between U-MTNN and W-MTNN models (we investigate
dataset-specific effects of per-task weighting in
Section~\ref{sec:dataset_size}).

\begin{table*}[tbp]
    \caption{Comparisons between neural network models. Differences between STNN,
    U-MTNN, and W-MTNN models with the same core (hidden layer) architecture are reported as
    median $\Delta$AUC values and sign test 95\% confidence intervals. Bold
    values indicate confidence intervals that do not include 0.5.}
    \label{table:mtnn_vs_stnn}
    \centering
    \rowcolors{2}{lightgray}{}
    \sisetup{detect-weight=true,detect-inline-weight=math}
    \begin{tabular}{ l l S c S c }
    \toprule
     & & \multicolumn{2}{c}{MTNN - STNN} &
         \multicolumn{2}{c}{W-MTNN - U-MTNN} \\
    \cmidrule(lr){3-4} \cmidrule(lr){5-6}
     & Model &
    {\makecell{Median \\ $\Delta$AUC}} & \makecell{Sign Test \\ 95\% CI} &
    {\makecell{Median \\ $\Delta$AUC}} & \makecell{Sign Test \\ 95\% CI} \\
    \midrule
    \cellcolor{white} & (1000) & 0.010 & (\num{0.43}, \num{0.80}) &  &  \\
    \cellcolor{white} & (4000) & 0.012 & (\num{0.43}, \num{0.80}) &  &  \\
    \cellcolor{white} & (2000, 100) & 0.015 & (\num{0.39}, \num{0.77}) &  &  \\
    \cellcolor{white} & (2000, 1000) & 0.026 & (\num{0.47}, \num{0.84}) &  &  \\
    \multirow{-5}{*}{\cellcolor{white} U-MTNN} & (4000, 2000, 1000, 1000) & 0.023 & (\num{0.43}, \num{0.80}) &  &  \\
    \midrule
    \cellcolor{white} & (1000) & \bfseries 0.017 & \bfseries (\num{0.52}, \num{0.87}) & 0.002 & (\num{0.37}, \num{0.76}) \\
    \cellcolor{white} & (4000) & 0.007 & (\num{0.47}, \num{0.84}) & 0.002 & (\num{0.35}, \num{0.73}) \\
    \cellcolor{white} & (2000, 100) & 0.004 & (\num{0.39}, \num{0.77}) & -0.002 & (\num{0.28}, \num{0.68}) \\
    \cellcolor{white} & (2000, 1000) & \bfseries 0.032 & \bfseries (\num{0.57}, \num{0.90}) & 0.005 & (\num{0.43}, \num{0.80}) \\
    \multirow{-5}{*}{\cellcolor{white} W-MTNN} & (4000, 2000, 1000, 1000) & 0.033 & (\num{0.43}, \num{0.80}) & 0.004 & (\num{0.43}, \num{0.80}) \\
    \bottomrule
    \end{tabular}
\end{table*}

\begin{figure}[tb]
  \centering
  \includegraphics[width=\linewidth]{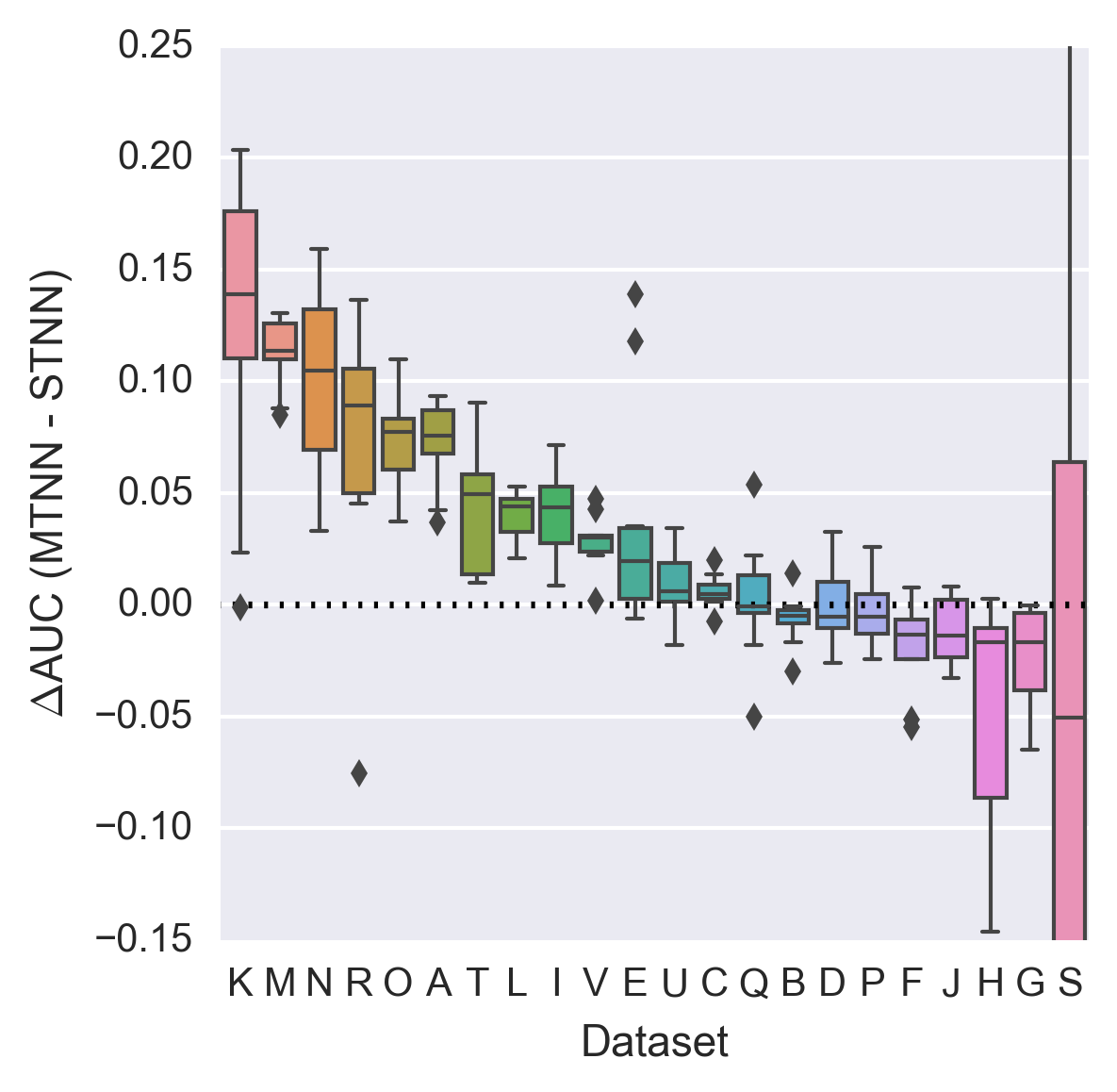}
  \caption{Box plots showing $\Delta$AUC values between MTNN
  and STNN models with the same core architecture. Each box plot
  summarizes 10 $\Delta$AUC values, one for each combination of model
  architecture (e.g.~(2000,~1000)) and task weighting strategy (U-MTNN or
  W-MTNN).}
  \label{fig:multitask_effect_box}
\end{figure}

Although multitask effects varied between models, the effects observed for
individual datasets were relatively independent of model architecture or task
weighting strategy~(\figurename~\ref{fig:multitask_effect_box}).

\tablename~\ref{appendix:table:architecture_comparison} shows pairwise
comparisons between model architectures in the same class (e.g.~STNN).
No architecture gave significant improvements over all others, suggesting that the effectiveness of
any given architecture is highly dependent on the available data. As such, we
cannot draw any general conclusions about the utility of one architecture over
another except to say that the (2000,~1000) W-MTNN architecture achieved the
highest median AUC for our datasets and consistently improved upon logistic
regression, random forest, and (2000,~1000) STNN models.

\subsection{Factors affecting multitask learning}

Several reports have shown that, on average, multitask learning improves
performance relative to single-task models~\citep{dahl2014multi, ma2015deep,
ramsundar2015massively, mayr2015deeptox}. However, these studies also include
examples of datasets that see only small or even negative effects in a multitask
setting. In this section, we investigate several factors that may contribute
directly or indirectly to observed multitask effects.

\subsubsection{Individual dataset size}
\label{sec:dataset_size}

\begin{figure}[tb]
  \centering
  \includegraphics[width=\linewidth]{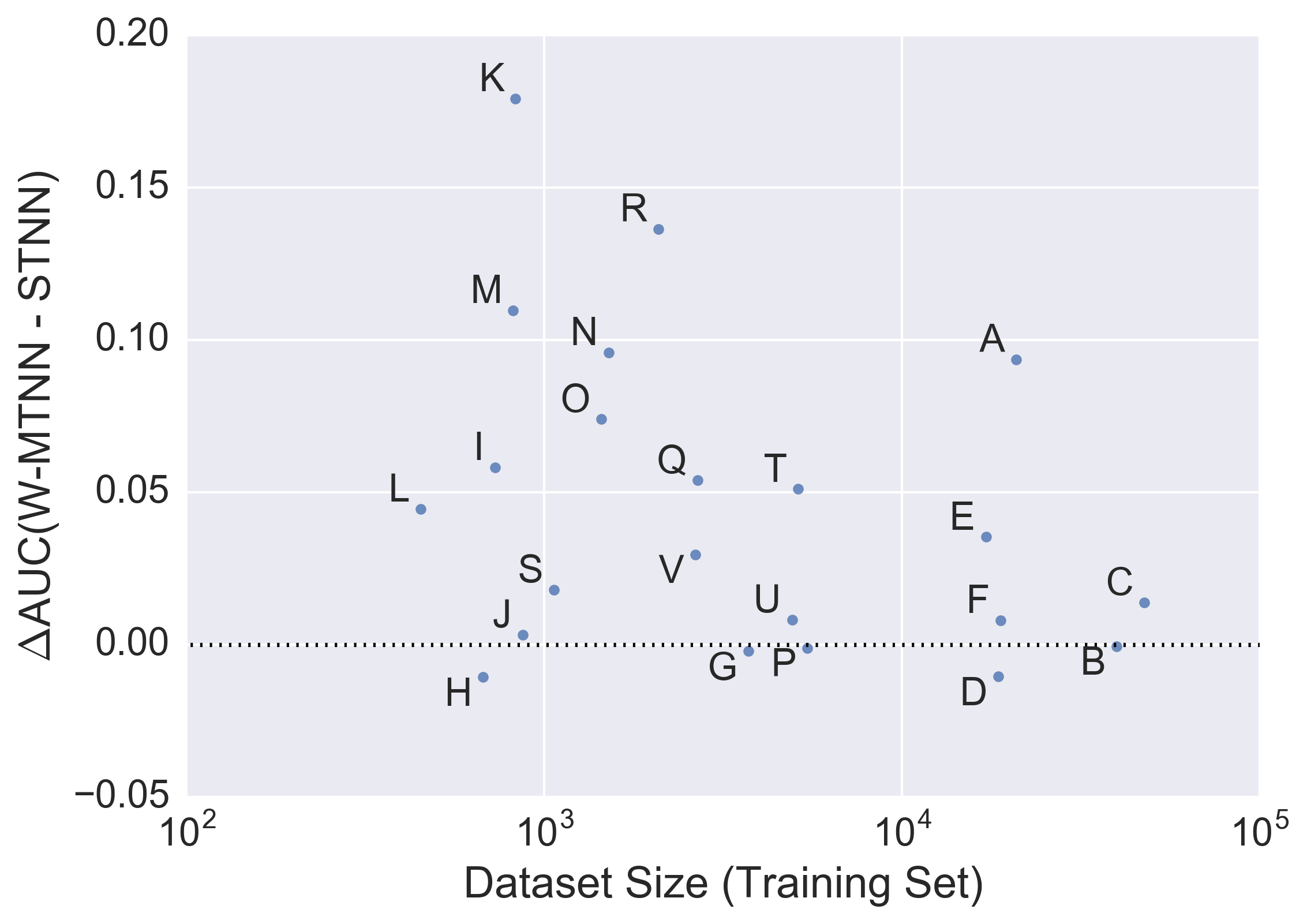}
  \caption{Multitask benefit as a function of dataset size (log scale, training set only; $r^2 \approx 0.12$). Multitask benefit
  for each task is calculated as the difference in AUC between the
  (2000,~1000)~W-MTNN and (2000,~1000)~STNN models.}
  \label{fig:dataset_size}
\end{figure}

It is possible that the size of individual datasets affects multitask
performance. \figurename~\ref{fig:dataset_size} shows a plot of multitask
benefit \emph{vs.} dataset size for the (2000,~1000)~W-MTNN from
\tablename~\ref{table:main_results}. There is a slight negative trend,
suggesting that larger datasets benefit less from multitask training, possibly
because they have enough data to generate a useful hidden layer representation
without additional side information. These results confirm previous
work showing multitask
improvement on relatively small datasets~\citep{dahl2014multi, ma2015deep, ramsundar2015massively} and should be encouraging since
pharmaceutical datasets are usually much smaller than those used for other deep
learning applications.

As an additional experiment with dataset size, we trained multitask models
for the ``small'' datasets in our collection with fewer than \num{10000} data
points (datasets G--V) to test whether small datasets were overwhelmed by larger tasks despite
the use of per-task weighting. \tablename~\ref{appendix:table:small_deltas}
shows comparisons between models trained using only small
datasets and full (22 task) multitask models for the 16 small datasets. We did not observe consistent
improvements when training only on small datasets, although some models gave
improvements for individual datasets~(\figurename~\ref{fig:appendix:subset_vs_full_small}).

We also considered the possibility that U-MTNN and W-MTNN models did not have
significantly different performance due to size-dependent effects; e.g.~task
weighting may improve performance on small datasets but hurt performance on
large datasets, producing a net effect that appears insignificant. However, we
did not observe a size-dependent trend when we compared U-MTNN and W-MTNN
performance on individual datasets~(\figurename~\ref{fig:appendix:dataset_size_weighted}).

\subsubsection{Total amount of data}
\label{sec:total_data}

Deep learning models have many parameters and perform best when they are trained
on huge amounts of data (typically millions of examples). The simplest neural
network models in our experiments have a single hidden layer with 1000 units,
requiring over 1~M learned parameters (not including the task-specific softmax
heads). In contrast, a multitask model trained on all of the datasets in our
collection has fewer than \num{200000} training set data points, which may
simply not be enough data to build effective models or control overfitting
(despite the use of dropout).

In the spirit of previous work that utilized large amounts of public data to
enhance multitask learning~\citep{ramsundar2015massively, unterthiner2014deep},
we constructed a side information dataset using internal Vertex data as well as
public sources (see Section~\ref{sec:appendix:datasets} for details). In
combination with our original datasets, this larger collection contained
approximately 38~M data points divided across more than 550 tasks. If multitask
effects can be modulated by the total amount of data and/or tasks, we
reasoned that this larger dataset collection should significantly improve the
average performance on the datasets in \tablename~\ref{table:datasets}. Training
each model on this larger dataset collection required several weeks with a
single GPU (all models were trained for at least 40~M steps). We note that some validation
set AUC scores were still increasing when training was stopped, so there is some
risk of underfitting in the results that follow; training curves for the tasks
in our dataset collection are shown in \figurename~\ref{fig:appendix:mtnn_training_curves}
and \figurename~\ref{fig:appendix:w_mtnn_training_curves}.

\tablename~\ref{appendix:table:public_results} reports performance on the
datasets from \tablename~\ref{table:datasets} for models trained with additional
side information. Notably, the (2000,~1000) and (4000,~2000,~1000,~1000) W-MTNN
models trained with side information achieved some of the highest median AUC
scores of all reported models and consistently improved upon STNN models with
the same core architecture~(\tablename~\ref{appendix:table:public_mtnn_vs_stnn}). Additionally, per-task
weighting led to significant performance improvements for most architectures.
However, adding side
information did not consistently improve performance relative to the MTNN
models reported above; comparisons between multitask models trained
with and without additional side information showed that, on average, adding
side information either had no significant effect or damaged performance~(\tablename~\ref{table:mtnn_vs_side}).

\begin{table*}[htb]
    \caption{Comparisons between multitask models trained with and without additional
    side information (SI). We report the median $\Delta$AUC and sign test 95\%
    confidence interval comparing MTNN+SI and MTNN models with the same core
    architecture and task weighting strategy. Bold values indicate confidence
    intervals that do not include 0.5.}
    \label{table:mtnn_vs_side}
    \centering
    \rowcolors{1}{}{lightgray}
    \sisetup{detect-all=true}
    \begin{tabular}{ l l S c }
    \toprule
     & & \multicolumn{2}{c}{MTNN+SI - MTNN} \\
    \cmidrule(lr){3-4}
     & Model &
    {\makecell{Median \\ $\Delta$AUC}} &
    \makecell{Sign Test \\ 95\% CI} \\
    \midrule
    \cellcolor{white} & (1000) & \bfseries \color{red} -0.035 & \bfseries \color{red} (\num{0.07}, \num{0.39}) \\
    \cellcolor{white} & (4000) & 0.004 & (\num{0.35}, \num{0.73}) \\
    \cellcolor{white} & (2000, 100) & \bfseries \color{red} -0.041 & \bfseries \color{red} (\num{0.10}, \num{0.43}) \\
    \cellcolor{white} & (2000, 1000) & -0.020 & (\num{0.16}, \num{0.53}) \\
    \multirow{-5}{*}{\cellcolor{white} U-MTNN} & (4000, 2000, 1000, 1000) & -0.027 & (\num{0.20}, \num{0.57}) \\
    \midrule
    \cellcolor{white} & (1000) & -0.009 & (\num{0.23}, \num{0.61}) \\
    \cellcolor{white} & (4000) & -0.000 & (\num{0.31}, \num{0.69}) \\
    \cellcolor{white} & (2000, 100) & 0.001 & (\num{0.31}, \num{0.69}) \\
    \cellcolor{white} & (2000, 1000) & 0.002 & (\num{0.31}, \num{0.69}) \\
    \multirow{-5}{*}{\cellcolor{white} W-MTNN} & (4000, 2000, 1000, 1000) & 0.007 & (\num{0.39}, \num{0.77}) \\
    \bottomrule
    \end{tabular}
\end{table*}

Inspection of per-dataset differences revealed that the effect of adding side
information was dataset-dependent~(\figurename~\ref{fig:appendix:mtnn_vs_side}). In contrast to the trend observed
in \figurename~\ref{fig:dataset_size}, some of the largest datasets in our
collection benefited the most from additional side information, including three
datasets (D, E, and F) with related targets.
These datasets had additional related assays in the included side information,
suggesting that these improvements might be due to additional related
training examples rather than an increase in the total amount of data.

\subsubsection{Task relatedness}
\label{sec:relatedness}

It is possible that the
datasets in our collection are diverse enough that na\"{i}vely combining them
into a single multitask model introduces competing training signals that dampen
multitask effects (see~\citet{caruana1997multitask} for a more thorough analysis
of multitask learning). To test this hypothesis, we trained multitask models using
subsets of the datasets in our collection that are related by similar targets
and are therefore most likely to benefit from multitask learning~(\tablename~\ref{appendix:table:subset_datasets}). We note that
\citet{erhan2006collaborative} also constructed models using subsets of related
tasks, using pairwise correlations between labels for shared compounds as a
measure of task similarity.

\tablename~\ref{appendix:table:subset_deltas} shows comparisons between subset
and full (22 task) multitask models for the 10 datasets used to train subset models.
These comparisons skew toward worse performance for subset models, but none of
the differences are statistically significant. Importantly, the subset models
had fewer training examples than full multitask models, and it is possible that
any additional multitask benefits due to more deliberate task selection were dampened by
training with less data.
Box plots summarizing per-dataset differences between subset and full models are
given in \figurename~\ref{fig:appendix:subset_vs_full_subset}.

These results support the inclusion of as much data and as many tasks as
possible when training multitask models, without regard to task relatedness. In
light of the seemingly contradictory result from Section~\ref{sec:total_data},
we suggest that multitask benefits increase with the addition of more data up to
an inflection point where additional data does not help or even begins to
degrade performance. This inflection point is likely to vary between datasets
and will depend on relatedness to the other tasks in the model.

\subsection{Information leakage in multitask networks}
\label{sec:leakage}

Multitask models have the potential for \emph{information leakage} across tasks,
which can lead to overly optimistic validation results even when using temporal
validation. Information leakage occurs when the training set for one task is
unrealistically related to the test set for another task. We use the word
``unrealistically'' to suggest that there are situations where relatedness is not
unfair, as might occur when starting a new screening campaign against a target
that is very similar to one that has been screened previously. (We note that
information leakage in multitask networks was treated briefly by
\citet{ramsundar2015massively} when considering the effect of random
cross-validation on closely related tasks.)

Consider a multitask model trained using a temporal validation scheme. If each
dataset is divided into training and test data independently, it is possible
that training data for one task was generated in the future relative to
the training data for a second task. This would allow the second task to benefit
from information that should not realistically be available to that task and
partially defeats the purpose of temporal validation. We refer to this as
``leaky'' validation. Random cross-validation is always leaky (and maximally
so), because there is no respect to past or future whatsoever. The
impact of information leakage is expected to be proportional to task
relatedness, such that more-related tasks should benefit more
from leaked information than less-related tasks.

\begin{figure}[tb]
  \centering
  \includegraphics[clip,trim=0 275 365 0,width=\linewidth]{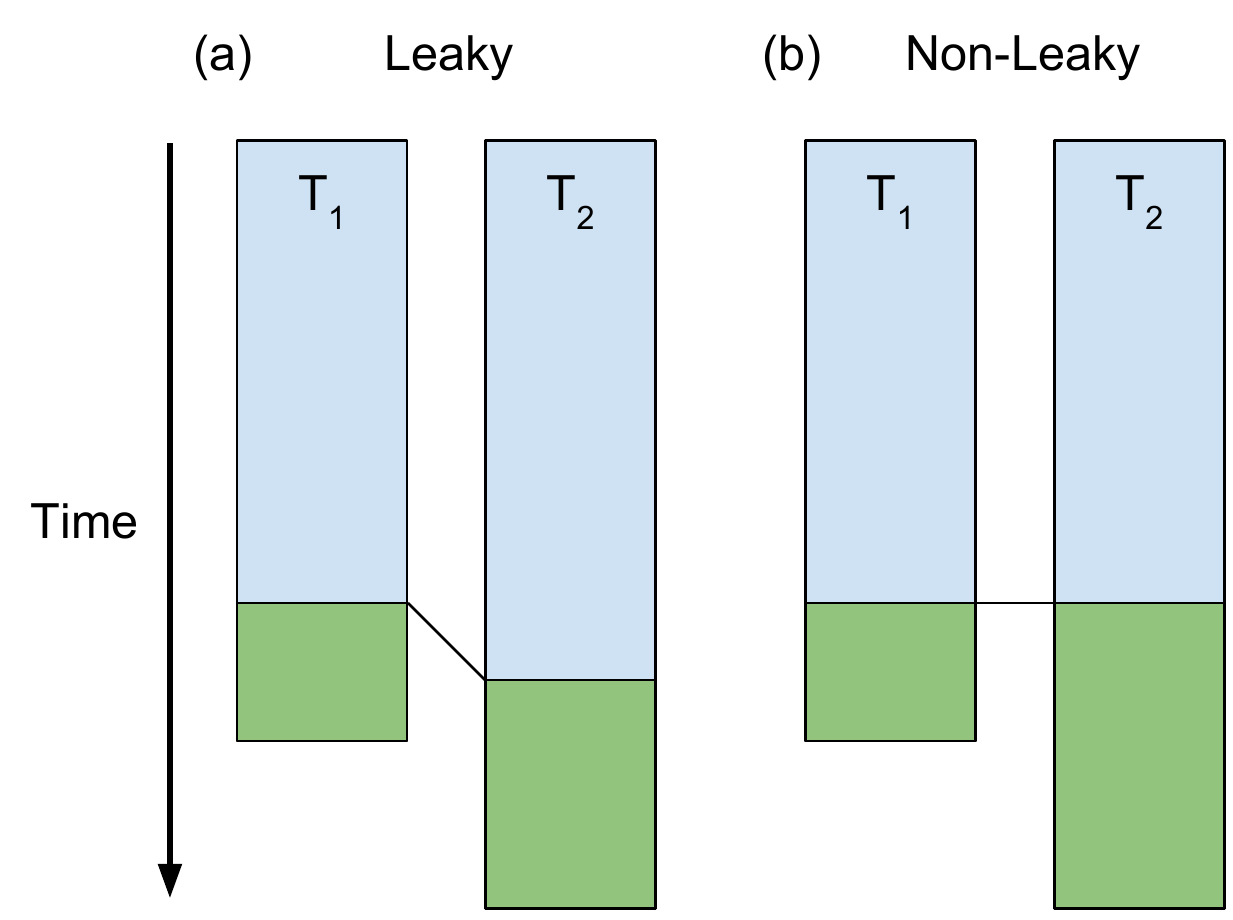}
  \caption{Leaky and non-leaky temporal validation of multitask models.
  In both strategies, datasets for two tasks (T$_1$ and T$_2$) are divided into
  training (blue) and test (green) data using specific time cutoffs (validation
  data is omitted for clarity). (a) Leaky validation divides each dataset
  independently, such that training data for one task could be in the future
  relative to training data for another task. (b) Non-leaky validation uses the
  same cutoff dates for all datasets.}
  \label{fig:validation_cartoon}
\end{figure}

Information leakage can be prevented by constructing models with a
definition of past and future that is consistent across all tasks,
i.e.~selecting a single point in time that
separates training and test data for all tasks. We refer to this strategy as
``non-leaky'' temporal validation. \figurename~\ref{fig:validation_cartoon} presents
graphical descriptions of leaky and non-leaky temporal validation.

We constructed non-leaky multitask models on our datasets for comparison with
the leaky multitask models presented above. Due to different date ranges
and dataset sizes, a single non-leaky model would give very poor statistics for
many tasks due to very small training or test sets. To correct for this issue,
and to make leaky and non-leaky models more directly comparable, we constructed
\emph{dataset-specific} non-leaky models. For each dataset in turn (the ``focus
dataset''), the training, validation, and test set cutoff dates used for
single-task and leaky multitask models were applied to the remaining datasets to
construct non-leaky multitask training, validation, and test sets, respectively.
As a result, the focus dataset in each non-leaky model had the same training,
validation, and test sets as its
single-task and leaky counterparts without including anachronistic side
information. We trained each non-leaky multitask model for
1~M steps and measured the test set performance for the focus dataset using the
training checkpoint that maximized performance on the validation set for the
focus task (data for the other tasks were used only as side information during
training).

\tablename~\ref{appendix:table:focused_results} reports the performance of the
non-leaky multitask models relative to random forest and logistic regression
baselines. Median AUC values for several models were quite different from the
leaky results in \tablename~\ref{table:main_results}, but the general trends in
performance were consistent for both validation strategies.
These results suggest that some information leakage occurred with leaky
validation on our dataset collection, leading to higher median AUC values for
some models (especially the (2000,~1000) and (4000,~2000,~1000,~1000)
architectures).
Comparisons between neural network models showed that some non-leaky models
outperformed their leaky counterparts relative to single-task models (compare
\tablename~\ref{table:mtnn_vs_stnn} and
\tablename~\ref{table:appendix:focused_mtnn_vs_stnn}), and per-task weighting
significantly improved performance for the (4000,~2000,~1000,~1000)
architecture. These results seem counterintuitive, but would not be unexpected
in a situation where the non-leaky cutoff dates assigned \emph{all} of the data
for one or more side information tasks to the training set. However, it is
unexpected that the changes in median AUC were not consistently in the same
direction, given that all non-leaky models were trained with the same data.
Direct comparisons between non-leaky and leaky models did not reveal any
significant differences in performance between the two validation
strategies~(\tablename~\ref{table:focused_vs_leaky}).

\begin{table*}[htb]
    \caption{Comparisons between leaky and non-leaky multitask models. We report the
    median $\Delta$AUC and sign test 95\% confidence interval comparing
    non-leaky and leaky models with the same core architecture and task
    weighting strategy. Bold values indicate confidence intervals that do not
    include 0.5.}
    \label{table:focused_vs_leaky}
    \centering
    \rowcolors{1}{}{lightgray}
    \sisetup{detect-weight=true,detect-inline-weight=math}
    \begin{tabular}{ l l S c }
    \toprule
     & & \multicolumn{2}{c}{Non-leaky - Leaky} \\
    \cmidrule(lr){3-4}
     & Model &
    {\makecell{Median \\ $\Delta$AUC}} &
    \makecell{Sign Test \\ 95\% CI} \\
    \midrule
    \cellcolor{white} & (1000) & 0.001 & (\num{0.35}, \num{0.73}) \\
    \cellcolor{white} & (4000) & 0.001 & (\num{0.35}, \num{0.73}) \\
    \cellcolor{white} & (2000, 100) & -0.002 & (\num{0.23}, \num{0.61}) \\
    \cellcolor{white} & (2000, 1000) & -0.010 & (\num{0.20}, \num{0.57}) \\
    \multirow{-5}{*}{\cellcolor{white} U-MTNN} & (4000, 2000, 1000, 1000) & -0.005 & (\num{0.20}, \num{0.57}) \\
    \midrule
    \cellcolor{white} & (1000) & 0.000 & (\num{0.31}, \num{0.69}) \\
    \cellcolor{white} & (4000) & -0.003 & (\num{0.23}, \num{0.61}) \\
    \cellcolor{white} & (2000, 100) & 0.002 & (\num{0.31}, \num{0.69}) \\
    \cellcolor{white} & (2000, 1000) & -0.011 & (\num{0.20}, \num{0.57}) \\
    \multirow{-5}{*}{\cellcolor{white} W-MTNN} & (4000, 2000, 1000, 1000) & 0.002 & (\num{0.35}, \num{0.73}) \\
    \bottomrule
    \end{tabular}
\end{table*}

We caution that these results are specific to our dataset collection and may not
hold for datasets with different date ranges, relatedness, or amounts of data.
In particular, the distribution of experiment dates for a dataset collection can
lead to very different amounts of training data for non-leaky and leaky models.
Our use of leaky validation for the majority of the analyses in this report
avoids sensitivity to the specific date ranges for our datasets---simplifying
the exploration of multitask effects---at the risk of unrealistic information
leakage.

\subsection{Temporal validation \emph{vs.} random cross-validation}

\citet{sheridan2013time} has shown that temporal validation gives more accurate
estimates of prospective model performance than random cross-validation in drug
discovery applications. Random cross-validation reduces
\emph{covariate shift}---the tendency for training and test examples to follow
different distributions. Covariate shift is especially common in pharmaceutical
data, since compounds are synthesized serially as project teams attempt to
optimize potency, off-target effects, and other molecular
properties~\citep{mcgaughey2016understanding}. In the worst case, compounds
used for training and test data in temporal validation could have entirely
different properties. By ignoring the temporal evolution of a dataset, random
cross-validation makes it much more likely that the training data is a good
approximation of the test data---a standard assumption in machine learning---but
a poor approximation of reality.

\begin{figure}[tb]
  \centering
  \includegraphics[width=\linewidth]{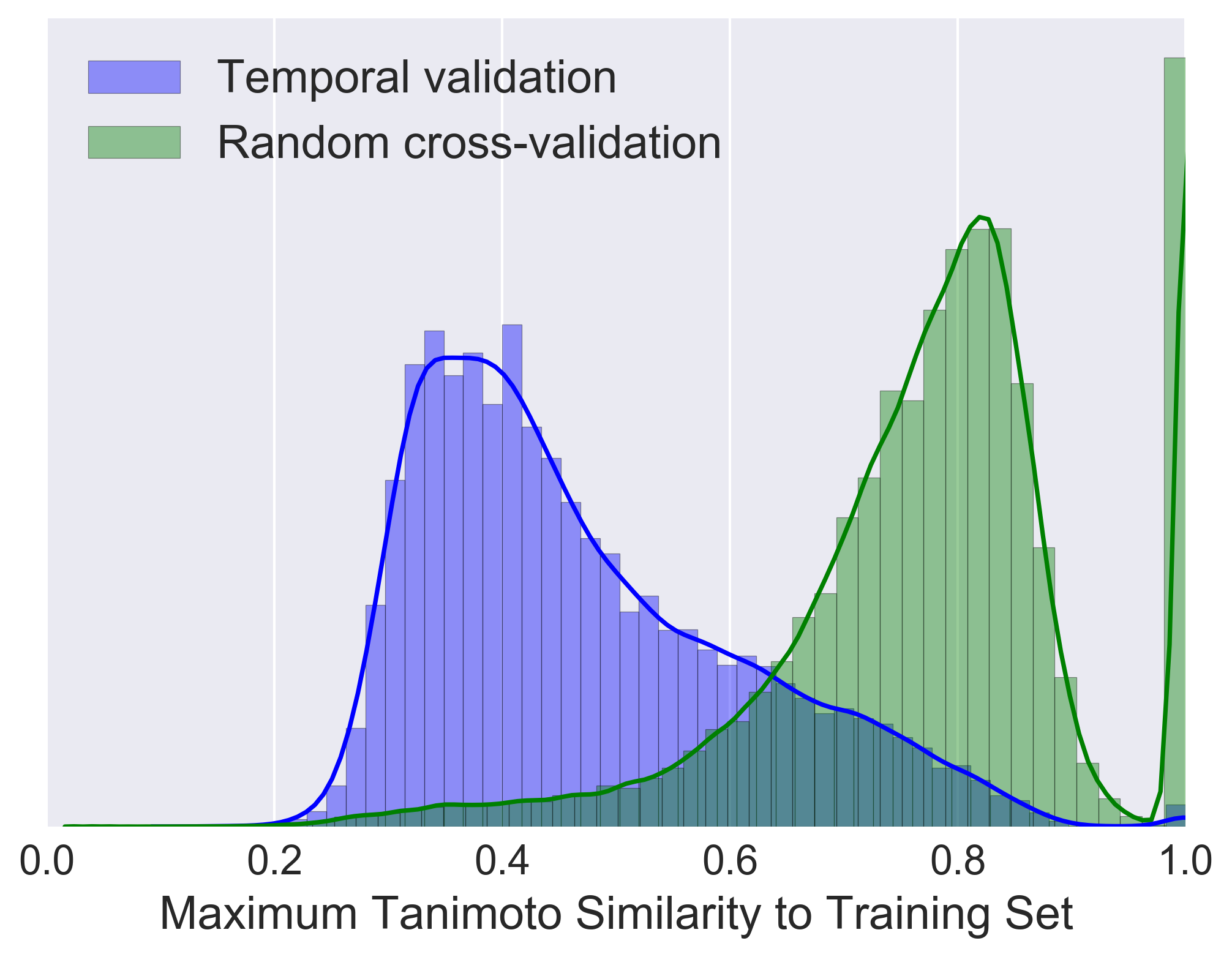}
  \caption{Distributions of maximum circular fingerprint Tanimoto similarity between each test
  compound and all training compounds (within each dataset) for temporal
  validation or random cross-validation. Note that the temporal validation
  histogram does not include similarities for validation set compounds.}
  \label{fig:covariate_shift}
\end{figure}

As a measure of covariate shift in our dataset collection, we measured the
maximum circular fingerprint Tanimoto similarity between each test compound and all training set
compounds in each dataset (note that temporal validation set compounds were excluded). Histograms of maximum similarities for temporal
validation and random cross-validation are shown in
\figurename~\ref{fig:covariate_shift}. As expected, test compounds were much
more likely to have a highly similar training compound when using random
cross-validation.

To compare the performance of models trained using temporal validation or random
cross-validation, we trained models using 5-fold random cross-validation and
measured 5-fold mean test set AUC values after
training for $\sim$1~M steps. \tablename~\ref{appendix:table:random_results}
shows that random cross-validation multitask models consistently improved upon logistic
regression and random forest baselines. Interestingly, the logistic regression
and random forest models also switched their relative performance compared to
temporal validation.
\tablename~\ref{appendix:table:random_mtnn_vs_stnn} shows that random cross-validation
multitask networks consistently beat their single-task counterparts regardless
of architecture or task weighting strategy. Overall,
multitask effects for random cross-validation models were more consistent than
for temporal models, although the effect sizes were generally
smaller (we attribute this to the fact that single-task random cross-validation
models achieved very high median AUC values that left little room for
improvement).

There were some differences between our temporal validation and random
cross-validation strategies that could lead to unfair comparisons; for instance,
random cross-validation models were trained on more data (80\% of each dataset \emph{vs.}
70\% for temporal models) and we did not use held-out validation sets to prevent
overfitting. To investigate the potential bias due to overfitting, we recalculated
performance metrics for random cross-validation models using an alternative evaluation strategy that maximized the
5-fold mean AUC for each task rather than evaluating after a fixed number of
training steps; these results are given in
Section~\ref{sec:appendix:random_cv_alternative}. This alternative evaluation
strategy revealed that random cross-validation models (especially STNNs) suffered from overfitting when
evaluated at $\sim$1~M steps, resulting in larger multitask effect sizes and
more biased sign test confidence intervals when comparing MTNNs with STNNs.
However, both strategies showed consistent multitask improvements relative to
logistic regression, random forest, and STNN models.

These results indicate that
multitask effects for specific models can vary depending on whether temporal
validation or random cross-validation is used to estimate model performance,
potentially leading to inconsistent conclusions regarding specific model
architectures.

\section{Discussion}

This report describes our efforts to apply multitask neural networks to industrial
datasets for ADMET targets at Vertex Pharmaceuticals. Our results confirm previous work demonstrating that
neural networks can outperform random forest and logistic regression baselines
and provide further evidence that multitask learning can improve performance
over single-task models.

Training models with large amounts of side information did not yield substantial
improvements relative to simpler multitask models; comparisons of
results on the 22 datasets in \tablename~\ref{table:datasets} did not show
consistent improvements for models trained with 500+ tasks \emph{vs.} 22 tasks,
although some individual tasks did see benefits from additional side
information. One possibility is that multitask learning on our datasets is
quickly subject to diminishing returns, reminiscent of the ``growth curves''
presented by~\citet{ramsundar2015massively}. Furthermore, the role of task
relatedness in multitask learning remains poorly understood, and it is possible
that our datasets are simply too dissimilar to the additional side information
to recapitulate the benefits seen in other studies.

Although our comparisons of temporal validation and random cross-validation were
not ideal, our experiments show that random cross-validation can lead to overly optimistic
estimates of multitask performance (especially regarding the relative performance of specific model
architectures). However, random cross-validation and temporal validation were
generally in agreement about broader trends in performance (e.g.~multitask
models can outperform single-task models in both validation paradigms).

Practically speaking, we have not identified any universally superior method or
architecture that will give optimal performance for every dataset. Accordingly,
we suggest that that recommended best practices should be treated as starting
points for building \emph{dataset-specific} models, especially in production
settings where it pays to have the best possible model for each individual task.

We expect that additional dataset curation and further
exploration of model architectures will yield modest improvements, but we are also
enthusiastic about recent work to improve the input representation for virtual
screening models. Recent work in this area includes
graph-based representations for deep learning systems such as ``neural graph
fingerprints''~\citep{duvenaud2015convolutional} and ``molecular graph
convolutions''~\citep{kearnes2016molecular}. Additional work on learning from
three-dimensional representations and handling conformational heterogeneity will
help to capture more of the relevant components of the systems and
processes we are attempting to model.

\section*{Acknowledgments}

We thank Patrick Riley, Kevin McCloskey, Georgia McGaughey, W. Patrick Walters,
Bharath Ramsundar, and the Vertex Modeling \& Informatics team for helpful
discussion. We also thank Gary Smith for computational support. S.K. was
supported by an internship at Vertex Pharmaceuticals Inc. and by Google Inc.
S.K. and V.P. also acknowledge support from from NIH 5U19AI109662-02.



\section*{Version information}

Submitted to the Journal of Computer-Aided Molecular Design.
Comments on arXiv versions:

\textbf{v2:} Corrected the summary in the Discussion of models trained with
additional side information.

\textbf{v3:} Updates in response to reviewer comments:
\begin{enumerate*}[label={\alph*)}]
\item deemphasized random cross-validation results due to non-ideal
    comparisons with temporal validation models,
\item added plots of validation AUC \emph{vs.} training step for models trained
    with side information,
\item updated table headers and figure axis labels for clarity,
\item additional minor changes to the main text and appendix.
\end{enumerate*}

\putbib
\end{bibunit}

\renewcommand{\thefigure}{A\arabic{figure}}
\renewcommand{\thetable}{A\arabic{table}}
\renewcommand\thesection{\Alph{section}}
\setcounter{section}{0}
\setcounter{figure}{0}
\setcounter{table}{0}

\let\Oldsection\section
\renewcommand{\section}{\FloatBarrier\Oldsection}
\let\Oldsubsection\subsection
\renewcommand{\subsection}{\FloatBarrier\Oldsubsection}
\let\Oldsubsubsection\subsubsection
\renewcommand{\subsubsection}{\FloatBarrier\Oldsubsubsection}

\onecolumn
\raggedbottom
\title{\textsc{Appendix}}
\author{}
\maketitle
\begin{bibunit}
\section{Appendix: Methods}
\subsection{Datasets}
\label{sec:appendix:datasets}

We included side information from proprietary and public sources. An
``internal'' collection of over 3000 additional Vertex assays with $\mu$M
readouts was downloaded from an internal database. These assays were further
divided by readout type---such as IC$_{50}$, EC$_{50}$ and $K_i$---yielding over
4000 datasets. Compounds with activities $\le 1$~$\mu$M were considered active,
and activities $\ge 10$~$\mu$M were considered inactive (intermediate activities
were discarded).

We used public data from PubChem BioAssay (PCBA)~\citep{wang2012pubchem} and the
enhanced directory of useful decoys (DUD-E)~\citep{mysinger2012directory},
similar to the approach taken by~\citet{ramsundar2015massively}. We downloaded
128 PCBA datasets and used their associated active/inactive annotations without
modification. All 102 DUD-E datasets were used, and each unique SMILES string
was treated as a separate example (regardless of compound identifier) since many
identifiers were associated with SMILES. For all datasets, multiple labels for
the same compound were merged if they agreed and discarded otherwise (for the
``internal'' datasets, multiple $\mu$M values were averaged before assigning an
activity class).

We employed a few simple filtering rules to limit the number of tasks in our
multitask models. All tasks were required to have at least 10 actives and 10
inactives in their training set. Additionally, side information datasets
(internal, PCBA, and DUD-E) were required to have at least 1000 training
examples.

For temporal models utilizing public side information, PCBA datasets were tagged
with their most recent modification date
(\tablename~\ref{appendix:table:pcba_dates}) and all the DUD-E datasets were
assigned to 14 July 2012 (the most recent modification date of the DUD-E
tarball). Note, however, that since these models used \emph{leaky} temporal
validation (see Section~\ref{sec:leakage}), these datasets were used in their
entirety as training data.

\begin{table*}[htb]
    \caption{Assay identification numbers (AIDs) and modification dates for PCBA
    datasets.}
    \label{appendix:table:pcba_dates}
    \centering
    \rowcolors{1}{}{lightgray}
    \begin{tabular}[t]{ l r | }
    AID & Modification Date \\
    \midrule
    411 & 2010-03-03 \\
    875 & 2010-07-12 \\
    881 & 2011-03-04 \\
    883 & 2010-07-06 \\
    884 & 2010-07-06 \\
    885 & 2010-07-06 \\
    887 & 2010-07-06 \\
    891 & 2010-07-06 \\
    899 & 2010-07-06 \\
    902 & 2010-07-06 \\
    903 & 2010-07-06 \\
    904 & 2010-07-06 \\
    912 & 2010-07-12 \\
    914 & 2010-07-06 \\
    915 & 2010-07-06 \\
    924 & 2010-07-06 \\
    925 & 2010-07-06 \\
    926 & 2010-07-06 \\
    927 & 2008-06-03 \\
    938 & 2010-07-06 \\
    995 & 2010-07-06 \\
    1030 & 2010-03-15 \\
    1379 & 2010-07-06 \\
    1452 & 2008-12-19 \\
    1454 & 2008-12-19 \\
    1457 & 2008-12-23 \\
    1458 & 2008-12-23 \\
    1460 & 2008-12-30 \\
    1461 & 2009-06-10 \\
    1468 & 2008-12-30 \\
    1469 & 2008-12-31 \\
    1471 & 2009-03-20 \\
    1479 & 2009-01-07 \\
    1631 & 2009-06-19 \\
    1634 & 2009-06-19 \\
    1688 & 2009-04-21 \\
    1721 & 2010-03-16 \\
    2100 & 2010-03-30 \\
    2101 & 2010-11-02 \\
    2147 & 2009-11-19 \\
    2242 & 2010-01-12 \\
    2326 & 2010-06-16 \\
    2451 & 2010-04-28 \\
    2517 & 2013-07-16 \\
    \end{tabular}
    \begin{tabular}[t]{ | l r | }
    AID & Modification Date \\
    \midrule
    2528 & 2010-03-15 \\
    2546 & 2010-03-13 \\
    2549 & 2010-03-15 \\
    2551 & 2010-03-15 \\
    2662 & 2010-09-28 \\
    2675 & 2014-02-22 \\
    2676 & 2010-03-25 \\
    463254 & 2010-09-23 \\
    485281 & 2010-09-27 \\
    485290 & 2010-09-28 \\
    485294 & 2010-09-29 \\
    485297 & 2010-09-29 \\
    485313 & 2010-10-04 \\
    485314 & 2010-09-29 \\
    485341 & 2010-10-04 \\
    485349 & 2010-10-04 \\
    485353 & 2010-11-12 \\
    485360 & 2010-10-05 \\
    485364 & 2010-10-06 \\
    485367 & 2010-11-05 \\
    492947 & 2010-11-29 \\
    493208 & 2011-02-14 \\
    504327 & 2011-02-22 \\
    504332 & 2011-02-22 \\
    504333 & 2011-02-22 \\
    504339 & 2011-02-22 \\
    504444 & 2011-03-15 \\
    504466 & 2011-03-16 \\
    504467 & 2011-03-16 \\
    504706 & 2011-04-27 \\
    504842 & 2011-06-23 \\
    504845 & 2011-06-23 \\
    504847 & 2011-06-24 \\
    504891 & 2012-05-22 \\
    540276 & 2011-11-18 \\
    540317 & 2011-07-28 \\
    588342 & 2011-09-07 \\
    588453 & 2011-10-05 \\
    588456 & 2011-10-06 \\
    588579 & 2011-10-20 \\
    588590 & 2011-10-20 \\
    588591 & 2011-10-20 \\
    588795 & 2011-11-16 \\
    588855 & 2011-12-06 \\
    \end{tabular}
    \begin{tabular}[t]{ | l r }
    AID & Modification Date \\
    \midrule
    602179 & 2012-04-05 \\
    602233 & 2012-01-27 \\
    602310 & 2012-11-27 \\
    602313 & 2012-03-08 \\
    602332 & 2012-03-13 \\
    624170 & 2012-05-22 \\
    624171 & 2012-05-22 \\
    624173 & 2012-05-30 \\
    624202 & 2012-05-24 \\
    624246 & 2012-06-01 \\
    624287 & 2012-06-12 \\
    624288 & 2013-12-14 \\
    624291 & 2012-06-13 \\
    624296 & 2012-06-15 \\
    624297 & 2012-06-15 \\
    624417 & 2012-07-26 \\
    651635 & 2012-10-18 \\
    651644 & 2012-10-16 \\
    651768 & 2012-11-16 \\
    651965 & 2013-01-03 \\
    652025 & 2013-02-12 \\
    652104 & 2013-03-14 \\
    652105 & 2013-03-14 \\
    652106 & 2013-03-14 \\
    686970 & 2013-05-09 \\
    686978 & 2013-05-16 \\
    686979 & 2013-05-16 \\
    720504 & 2013-07-09 \\
    720532 & 2013-07-20 \\
    720542 & 2013-07-27 \\
    720551 & 2013-07-31 \\
    720553 & 2013-07-31 \\
    720579 & 2013-08-21 \\
    720580 & 2013-08-19 \\
    720707 & 2013-10-31 \\
    720708 & 2013-10-31 \\
    720709 & 2013-10-31 \\
    720711 & 2013-10-31 \\
    743255 & 2014-01-16 \\
    743266 & 2014-01-30 \\
    \end{tabular}
\end{table*}

\section{Appendix: Results}
\subsection{Model performance}

\begin{table*}[htb]
    \caption{Pairwise comparisons between neural network model architectures.
    For each pair of models within a model class (e.g.~STNN), we report
    the median $\Delta$AUC and sign test 95\% confidence interval. Bold values
    indicate confidence intervals that do not include 0.5.}
    \label{appendix:table:architecture_comparison}
    \centering
    \rowcolors{2}{}{lightgray}
    \sisetup{detect-weight=true,detect-inline-weight=math}
    \begin{tabular}{ l l l S c }
    \toprule
     & & & \multicolumn{2}{c}{Model B - Model A} \\
    \cmidrule(lr){4-5}
     & Model A & Model B &
    {\makecell{Median \\ $\Delta$AUC}} &
    \makecell{Sign Test \\ 95\% CI} \\
    \midrule
    \cellcolor{white} & (1000) & (4000) & \bfseries 0.006 & \bfseries (\num{0.61}, \num{0.93}) \\
    \cellcolor{white} & (1000) & (2000, 100) & -0.004 & (\num{0.23}, \num{0.61}) \\
    \cellcolor{white} & (1000) & (2000, 1000) & -0.004 & (\num{0.20}, \num{0.57}) \\
    \cellcolor{white} & (1000) & (4000, 2000, 1000, 1000) & -0.004 & (\num{0.27}, \num{0.65}) \\
    \cellcolor{white} & (4000) & (2000, 100) & \bfseries \color{red} -0.006 & \bfseries \color{red} (\num{0.13}, \num{0.48}) \\
    \cellcolor{white} & (4000) & (2000, 1000) & -0.006 & (\num{0.20}, \num{0.57}) \\
    \cellcolor{white} & (4000) & (4000, 2000, 1000, 1000) & \bfseries \color{red} -0.011 & \bfseries \color{red} (\num{0.13}, \num{0.48}) \\
    \cellcolor{white} & (2000, 100) & (2000, 1000) & 0.002 & (\num{0.35}, \num{0.73}) \\
    \cellcolor{white} & (2000, 100) & (4000, 2000, 1000, 1000) & -0.001 & (\num{0.28}, \num{0.68}) \\
    \multirow{-10}{*}{\cellcolor{white} STNN} & (2000, 1000) & (4000, 2000, 1000, 1000) & -0.006 & (\num{0.16}, \num{0.53}) \\
    \midrule
    \cellcolor{white} & (1000) & (4000) & -0.002 & (\num{0.31}, \num{0.69}) \\
    \cellcolor{white} & (1000) & (2000, 100) & -0.005 & (\num{0.27}, \num{0.65}) \\
    \cellcolor{white} & (1000) & (2000, 1000) & \bfseries 0.012 & \bfseries (\num{0.52}, \num{0.87}) \\
    \cellcolor{white} & (1000) & (4000, 2000, 1000, 1000) & -0.004 & (\num{0.23}, \num{0.61}) \\
    \cellcolor{white} & (4000) & (2000, 100) & -0.009 & (\num{0.16}, \num{0.53}) \\
    \cellcolor{white} & (4000) & (2000, 1000) & 0.006 & (\num{0.43}, \num{0.80}) \\
    \cellcolor{white} & (4000) & (4000, 2000, 1000, 1000) & -0.005 & (\num{0.27}, \num{0.65}) \\
    \cellcolor{white} & (2000, 100) & (2000, 1000) & \bfseries 0.012 & \bfseries (\num{0.52}, \num{0.87}) \\
    \cellcolor{white} & (2000, 100) & (4000, 2000, 1000, 1000) & \bfseries 0.011 & \bfseries (\num{0.52}, \num{0.87}) \\
    \multirow{-10}{*}{\cellcolor{white} U-MTNN} & (2000, 1000) & (4000, 2000, 1000, 1000) & -0.009 & (\num{0.16}, \num{0.53}) \\
    \midrule
    \cellcolor{white} & (1000) & (4000) & -0.001 & (\num{0.27}, \num{0.65}) \\
    \cellcolor{white} & (1000) & (2000, 100) & \bfseries \color{red} -0.016 & \bfseries \color{red} (\num{0.13}, \num{0.48}) \\
    \cellcolor{white} & (1000) & (2000, 1000) & \bfseries 0.007 & \bfseries (\num{0.52}, \num{0.87}) \\
    \cellcolor{white} & (1000) & (4000, 2000, 1000, 1000) & -0.007 & (\num{0.23}, \num{0.61}) \\
    \cellcolor{white} & (4000) & (2000, 100) & -0.009 & (\num{0.23}, \num{0.61}) \\
    \cellcolor{white} & (4000) & (2000, 1000) & 0.005 & (\num{0.47}, \num{0.84}) \\
    \cellcolor{white} & (4000) & (4000, 2000, 1000, 1000) & -0.005 & (\num{0.23}, \num{0.61}) \\
    \cellcolor{white} & (2000, 100) & (2000, 1000) & \bfseries 0.020 & \bfseries (\num{0.57}, \num{0.90}) \\
    \cellcolor{white} & (2000, 100) & (4000, 2000, 1000, 1000) & \bfseries 0.016 & \bfseries (\num{0.57}, \num{0.90}) \\
    \multirow{-10}{*}{\cellcolor{white} W-MTNN} & (2000, 1000) & (4000, 2000, 1000, 1000) & -0.008 & (\num{0.16}, \num{0.53}) \\
    \bottomrule
    \end{tabular}
\end{table*}

\subsection{Factors affecting multitask learning}

\subsubsection{Individual dataset size}

\begin{table*}[htb]
    \caption{Comparisons between ``subset'' models trained only on small datasets
    (<\num{10000} data points) and ``full'' multitask models trained on all 22
    datasets in \tablename~\ref{table:datasets}. We report the median
    $\Delta$AUC and sign test 95\% confidence interval comparing subset and full
    multitask models with the same core architecture and task weighting
    strategy. Differences were calculated only for the 16 datasets that were used to
    build subset models. Bold values indicate confidence intervals that do
    not include 0.5.}
    \label{appendix:table:small_deltas}
    \centering
    \rowcolors{1}{}{lightgray}
    \sisetup{detect-all=true}
    \begin{tabular}{ l l S c }
    \toprule
     & & \multicolumn{2}{c}{Subset MTNN - Full MTNN} \\
    \cmidrule(lr){3-4}
     & Model &
    {\makecell{Median \\ $\Delta$AUC}} &
    \makecell{Sign Test \\ 95\% CI} \\
    \midrule
    \cellcolor{white} & (1000) & 0.005 & (\num{0.44}, \num{0.86}) \\
    \cellcolor{white} & (4000) & -0.003 & (\num{0.23}, \num{0.67}) \\
    \cellcolor{white} & (2000, 100) & -0.006 & (\num{0.18}, \num{0.61}) \\
    \cellcolor{white} & (2000, 1000) & -0.010 & (\num{0.14}, \num{0.56}) \\
    \multirow{-5}{*}{\cellcolor{white} U-MTNN} & (4000, 2000, 1000, 1000) & \bfseries \color{red} -0.014 & \bfseries \color{red} (\num{0.10}, \num{0.49}) \\
    \midrule
    \cellcolor{white} & (1000) & -0.007 & (\num{0.14}, \num{0.56}) \\
    \cellcolor{white} & (4000) & -0.003 & (\num{0.18}, \num{0.61}) \\
    \cellcolor{white} & (2000, 100) & -0.009 & (\num{0.23}, \num{0.67}) \\
    \cellcolor{white} & (2000, 1000) & -0.009 & (\num{0.14}, \num{0.56}) \\
    \multirow{-5}{*}{\cellcolor{white} W-MTNN} & (4000, 2000, 1000, 1000) & \bfseries \color{red} -0.007 & \bfseries \color{red} (\num{0.10}, \num{0.49}) \\
    \bottomrule
    \end{tabular}
\end{table*}

\begin{figure}[htbp]
  \centering
  \includegraphics[width=0.5\linewidth]{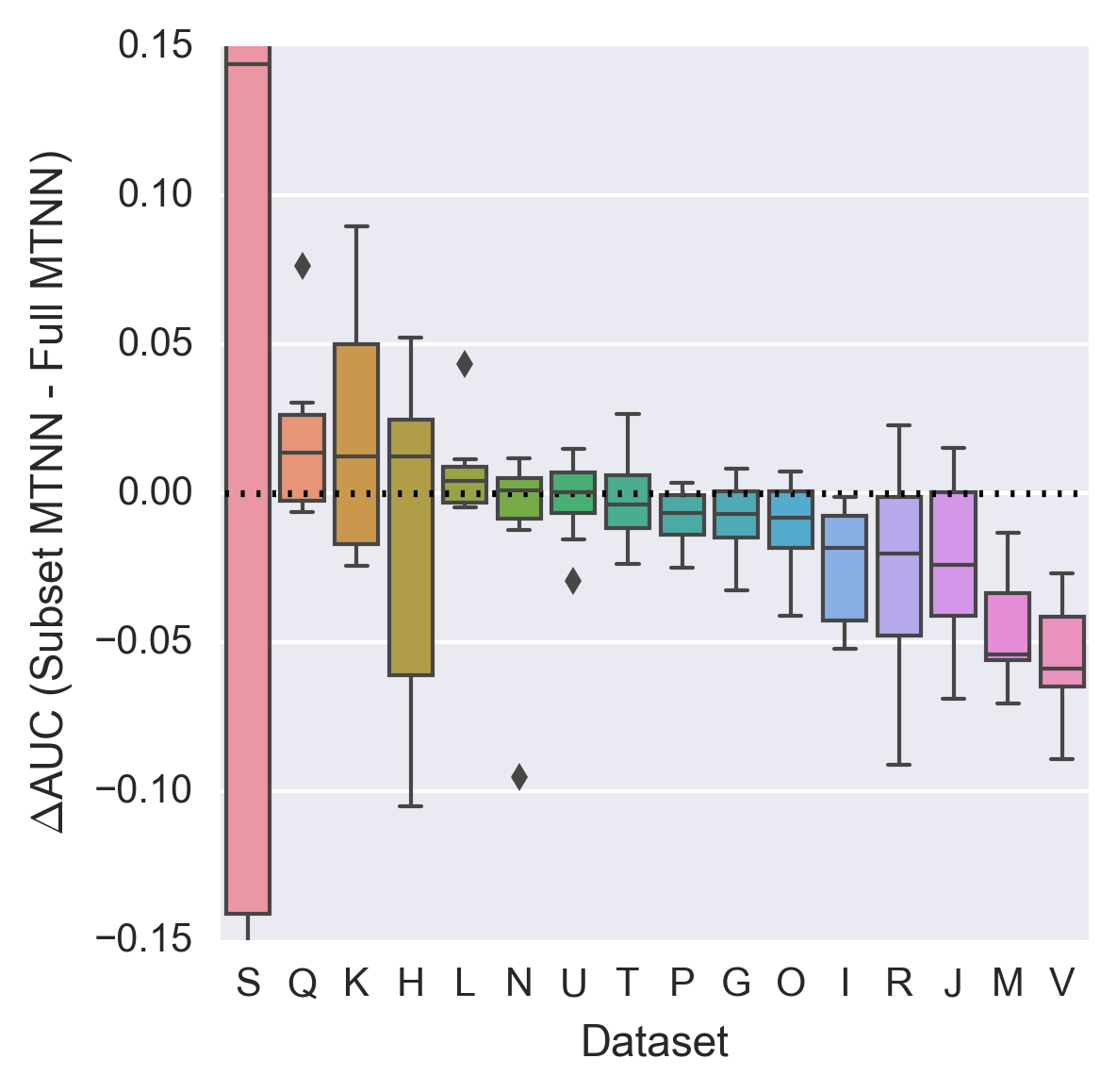}
  \caption{Box plots showing $\Delta$AUC values between ``subset'' MTNN models
  trained only on small datasets (<\num{10000} data points) and full MTNN models with
  the same core architecture. Each box plot summarizes 10 $\Delta$AUC values,
  one for each combination of model architecture (e.g.~(2000,~1000)) and task
  weighting strategy (U-MTNN or W-MTNN).}
  \label{fig:appendix:subset_vs_full_small}
\end{figure}

\begin{figure}[htbp]
  \centering
  \includegraphics[width=0.5\linewidth]{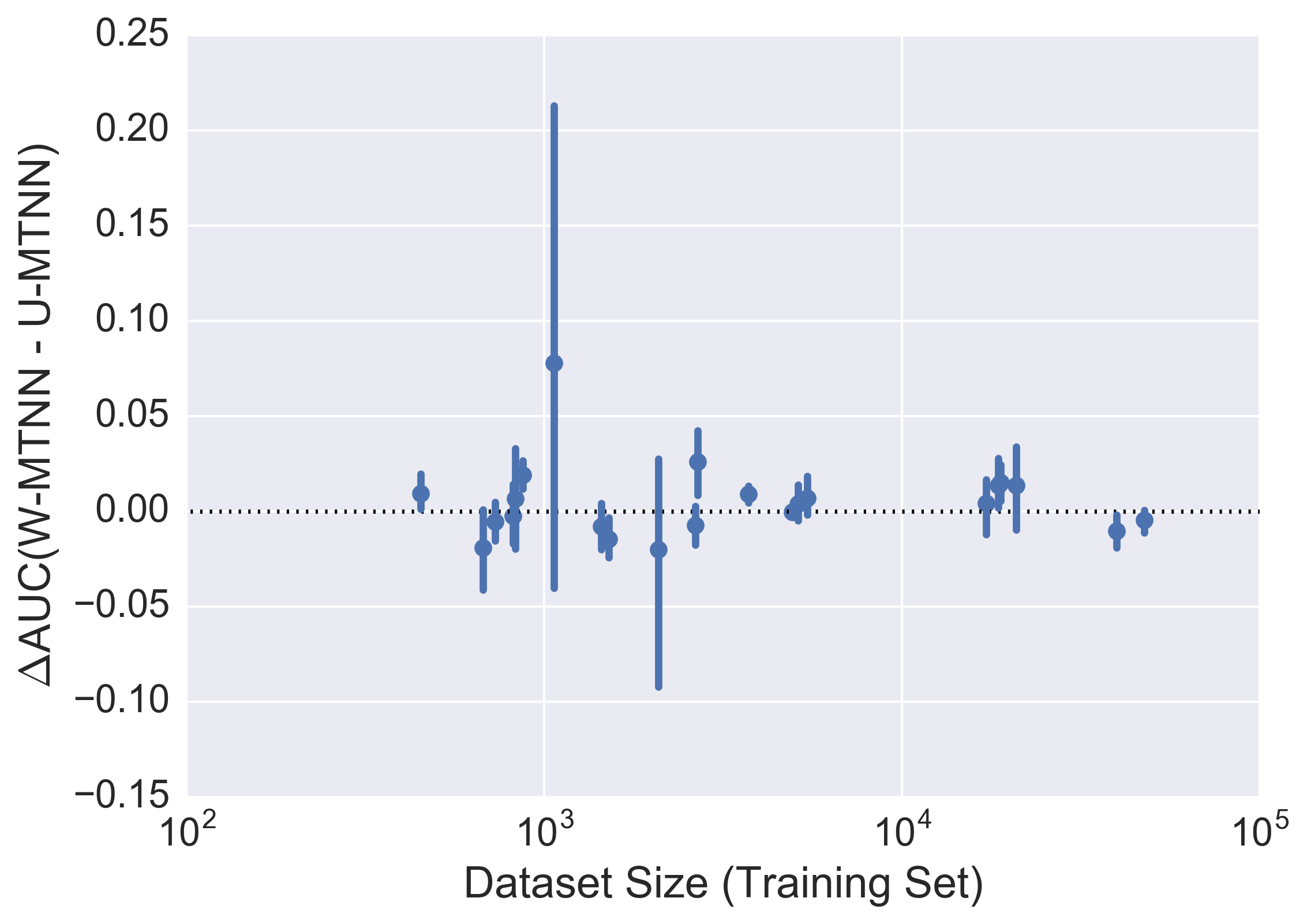}
  \caption{Differences between W-MTNN and U-MTNN test set AUC values for models
  with the same core architecture as a function of dataset size. Each point is
  the mean difference across all architectures; 95\% confidence intervals were
  calculated by bootstrapping.}
  \label{fig:appendix:dataset_size_weighted}
\end{figure}

\subsubsection{Total amount of data}
\label{sec:appendix:total_data}

\begin{figure}[htbp]
  \centering
  \begin{subfigure}{0.49\linewidth}
  \includegraphics[width=\linewidth]{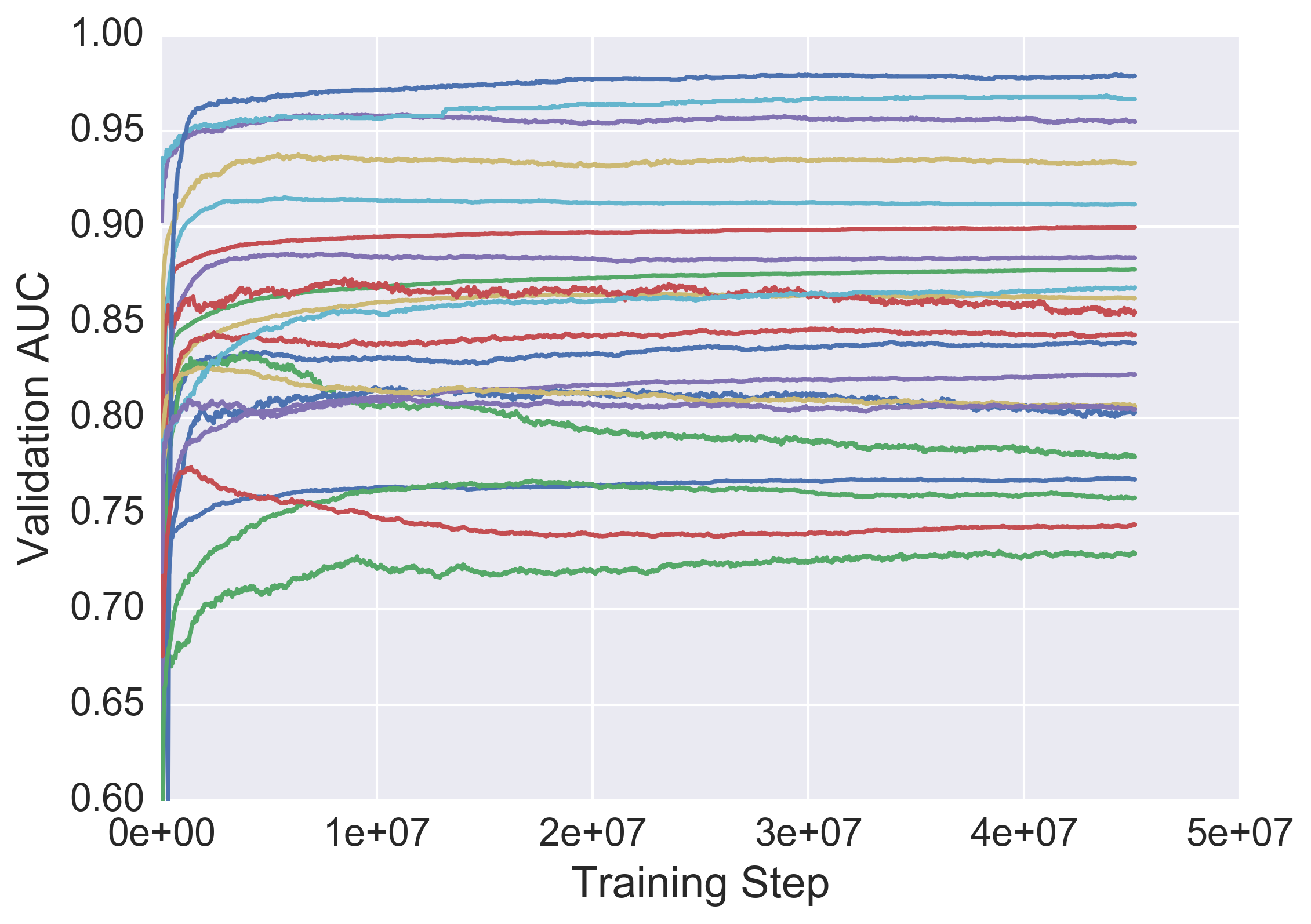}
  \caption{(1000) U-MTNN+SI}
  \end{subfigure}
  \begin{subfigure}{0.49\linewidth}
  \includegraphics[width=\linewidth]{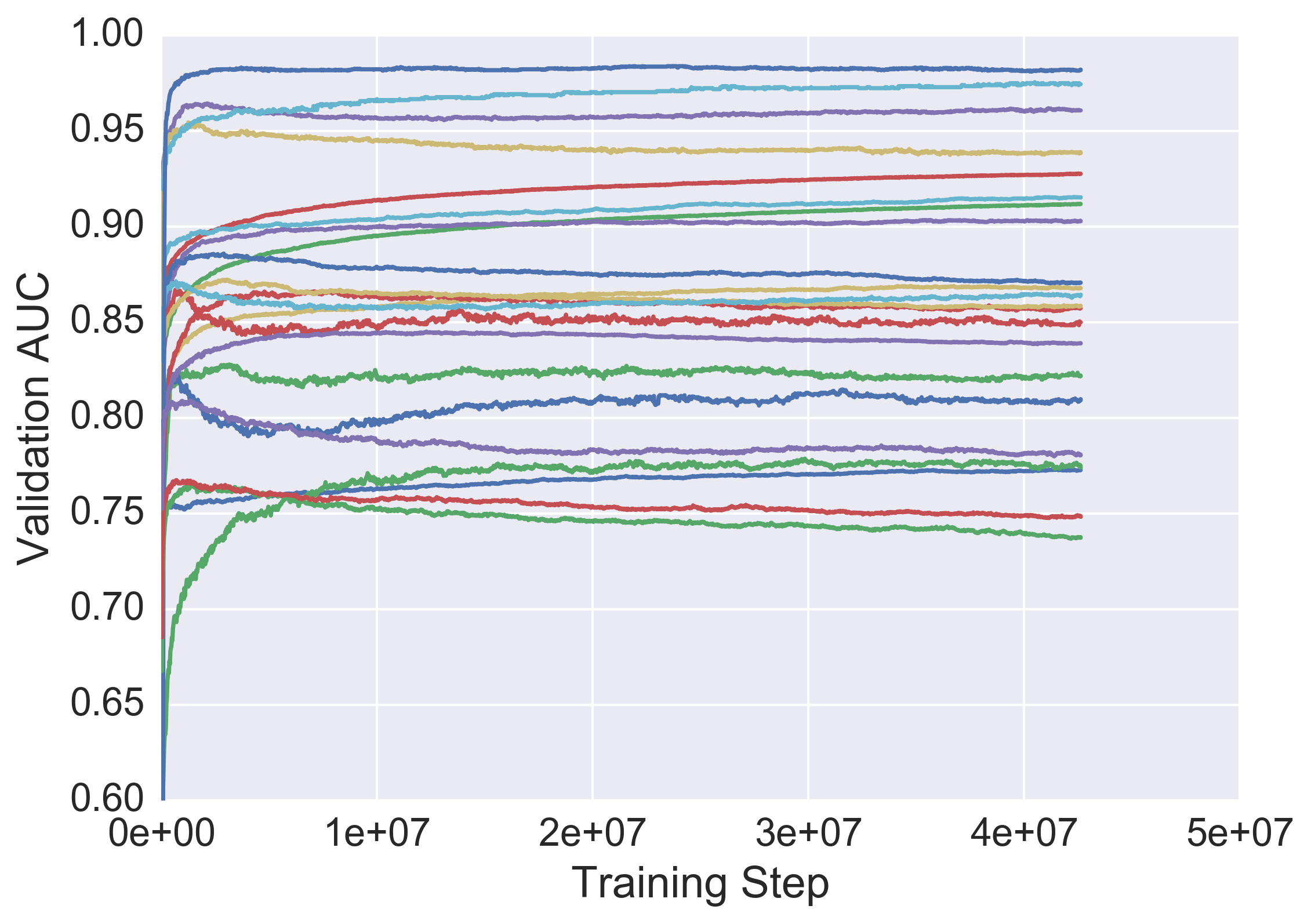}
  \caption{(4000) U-MTNN+SI}
  \end{subfigure}
  \begin{subfigure}{0.49\linewidth}
  \includegraphics[width=\linewidth]{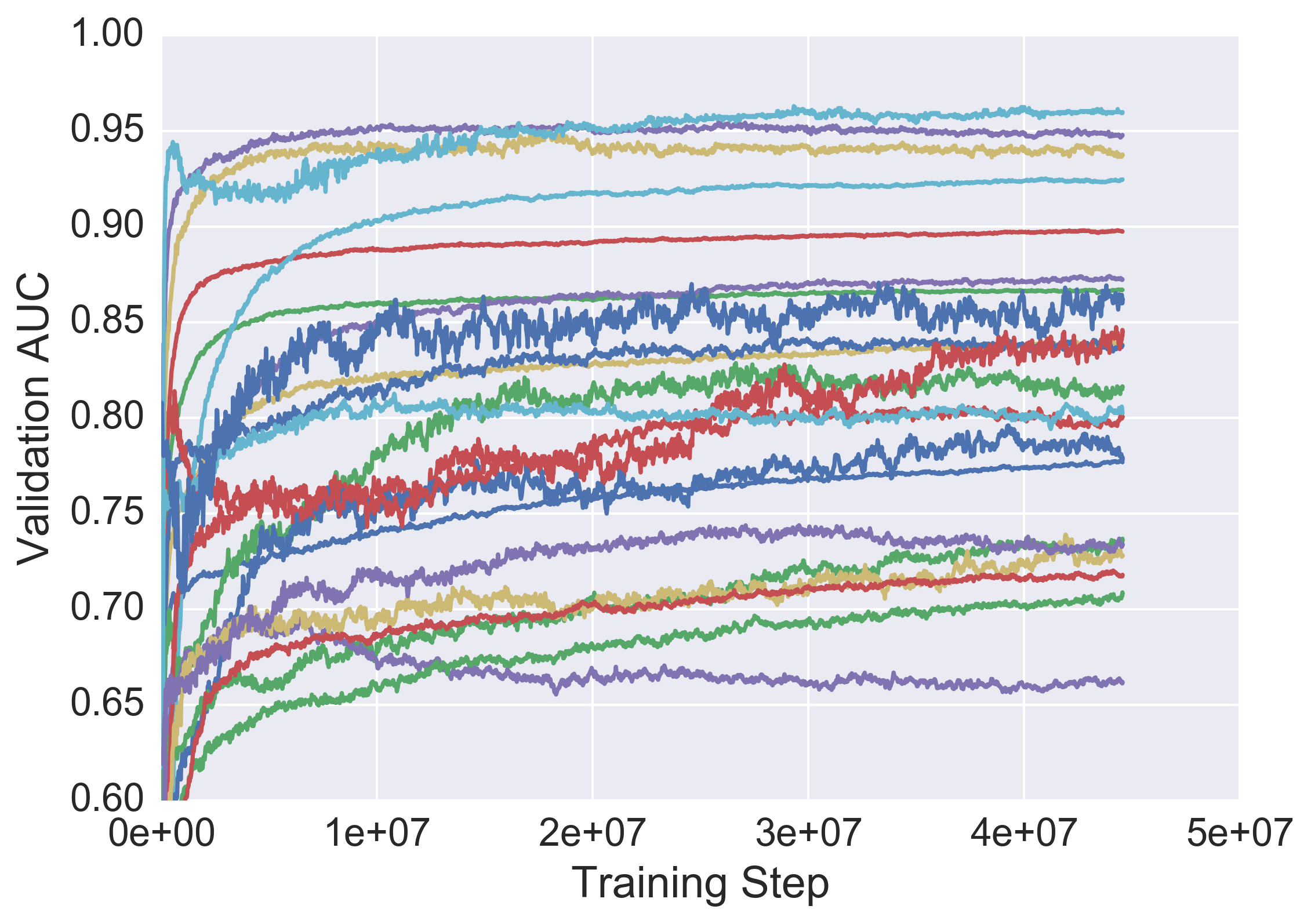}
  \caption{(2000,~100) U-MTNN+SI}
  \end{subfigure}
  \begin{subfigure}{0.49\linewidth}
  \includegraphics[width=\linewidth]{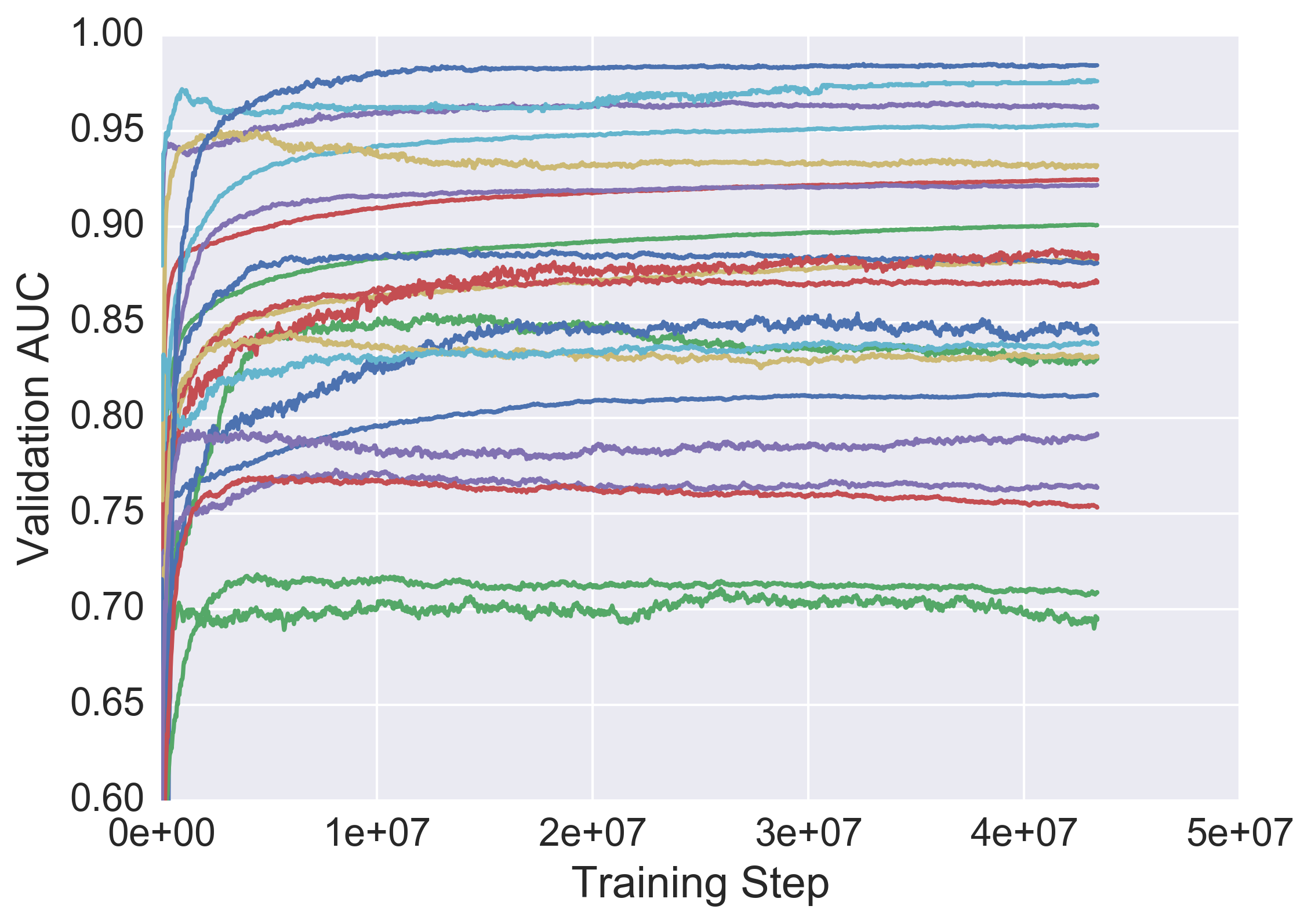}
  \caption{(2000,~1000) U-MTNN+SI}
  \end{subfigure}
  \begin{subfigure}{0.49\linewidth}
  \includegraphics[width=\linewidth]{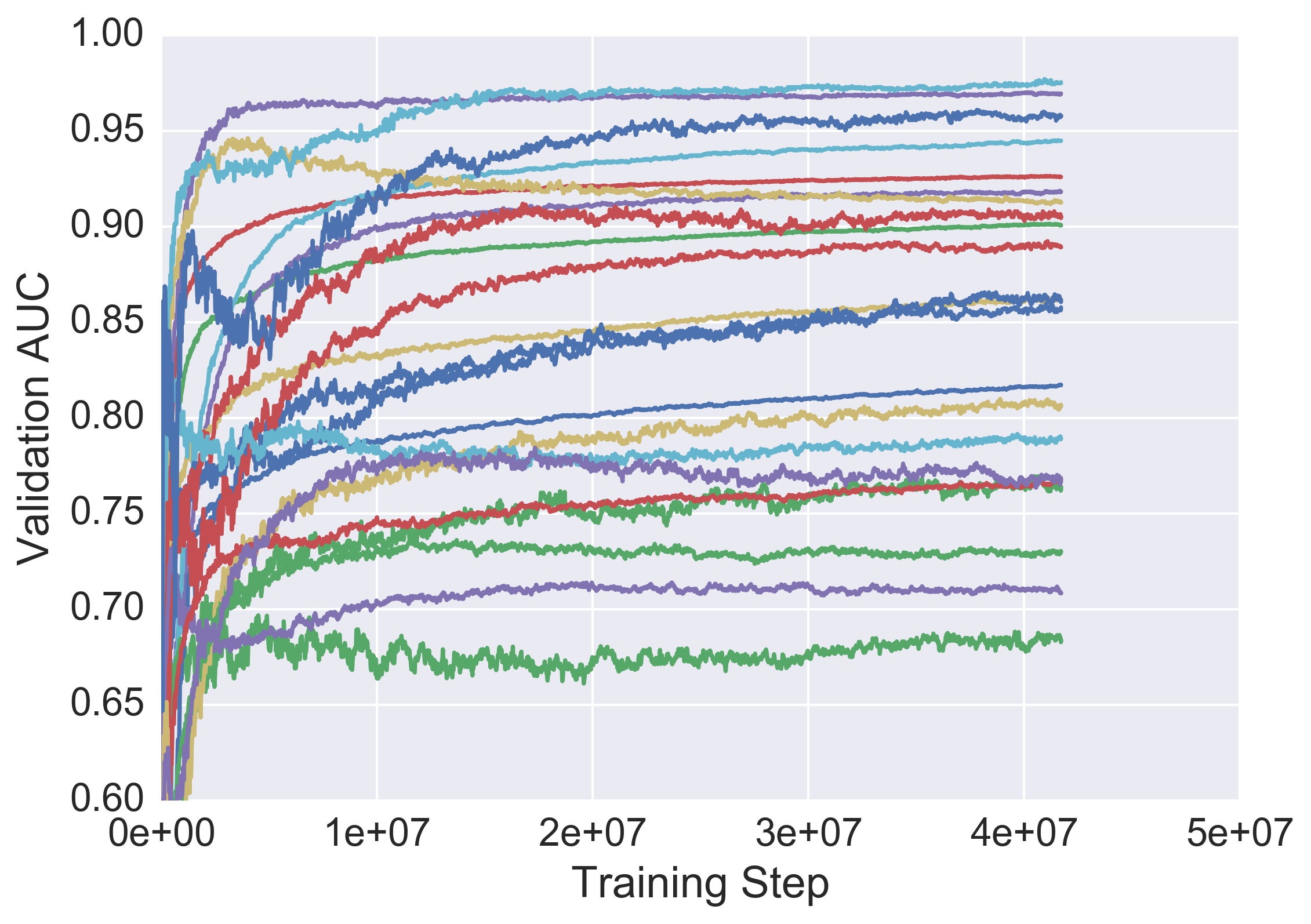}
  \caption{(4000,~2000,~1000,~1000) U-MTNN+SI}
  \end{subfigure}
  \caption{Validation set AUC for each task in multitask models trained with
    side information (U-MTNN+SI) as a function of training step.}
  \label{fig:appendix:mtnn_training_curves}
\end{figure}

\begin{figure}[htbp]
  \centering
  \begin{subfigure}{0.49\linewidth}
  \includegraphics[width=\linewidth]{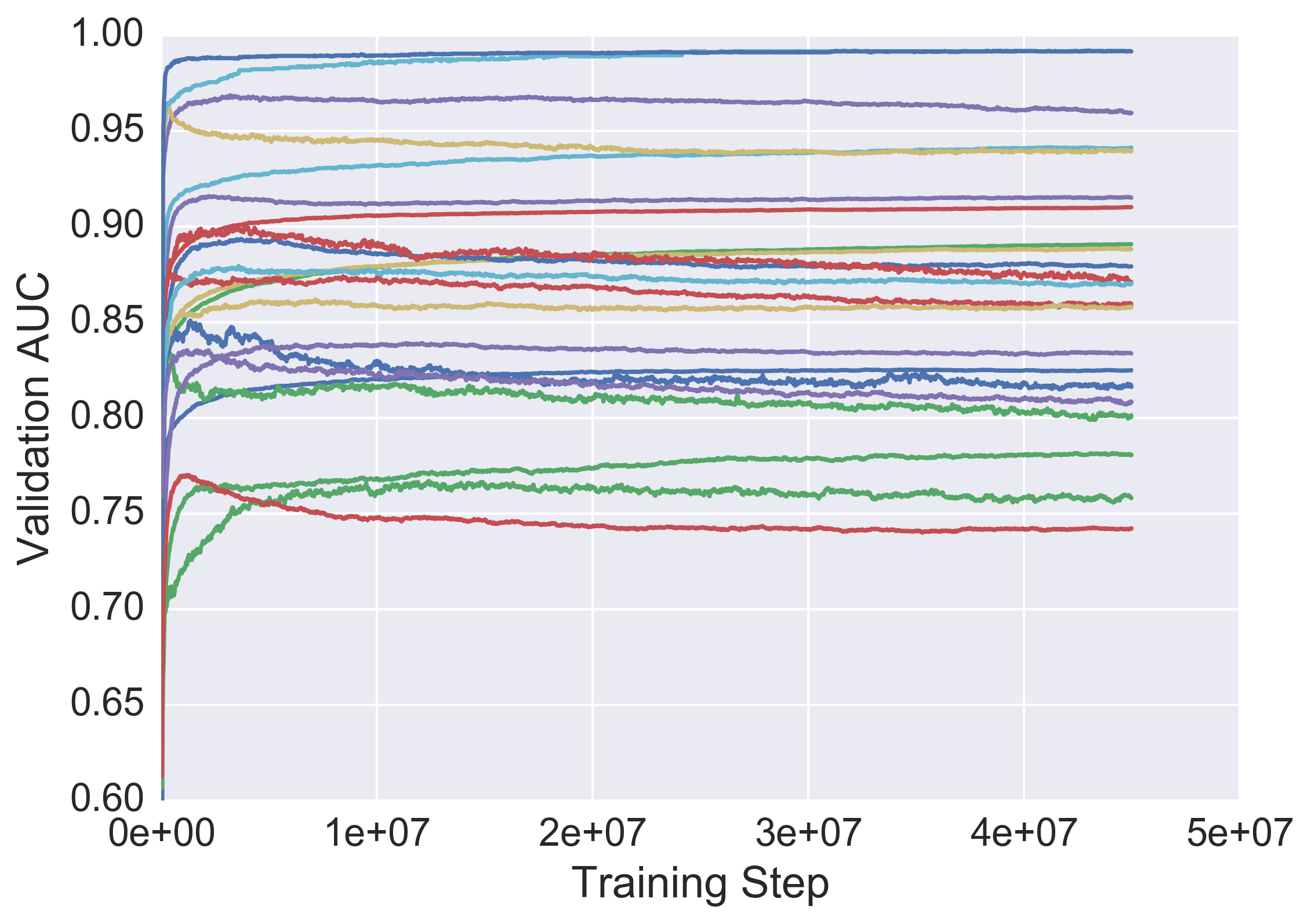}
  \caption{(1000) W-MTNN+SI}
  \end{subfigure}
  \begin{subfigure}{0.49\linewidth}
  \includegraphics[width=\linewidth]{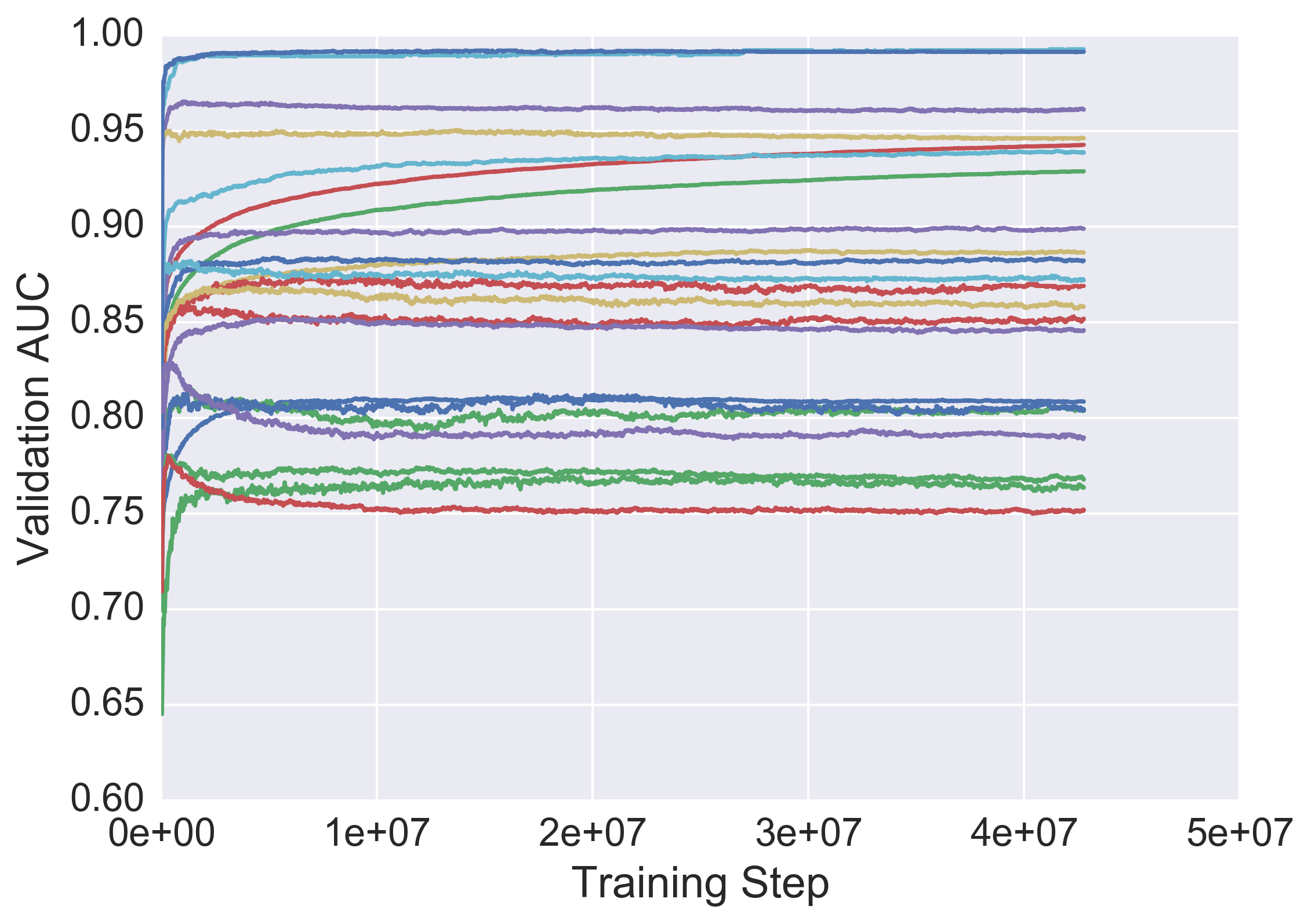}
  \caption{(4000) W-MTNN+SI}
  \end{subfigure}
  \begin{subfigure}{0.49\linewidth}
  \includegraphics[width=\linewidth]{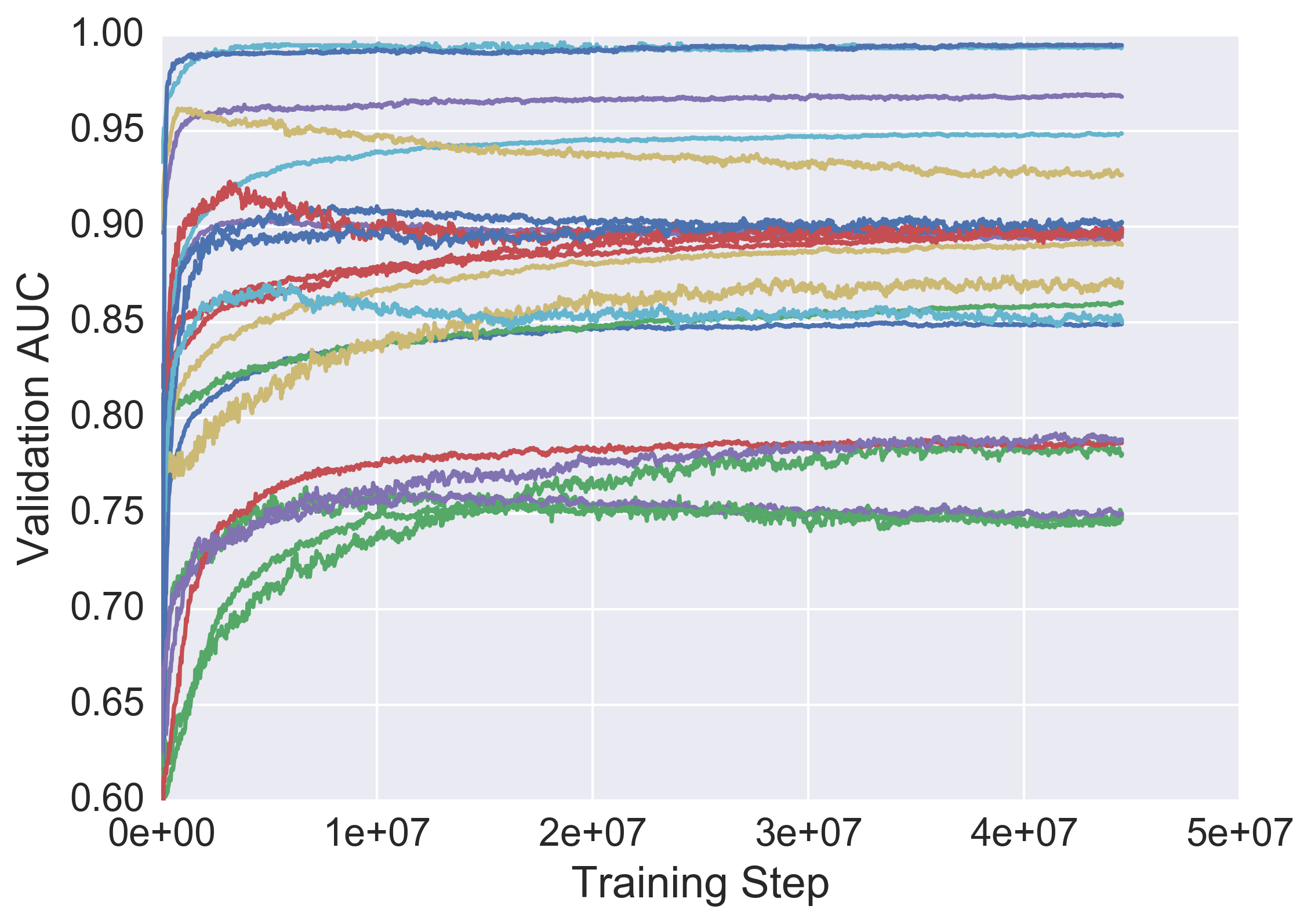}
  \caption{(2000,~100) W-MTNN+SI}
  \end{subfigure}
  \begin{subfigure}{0.49\linewidth}
  \includegraphics[width=\linewidth]{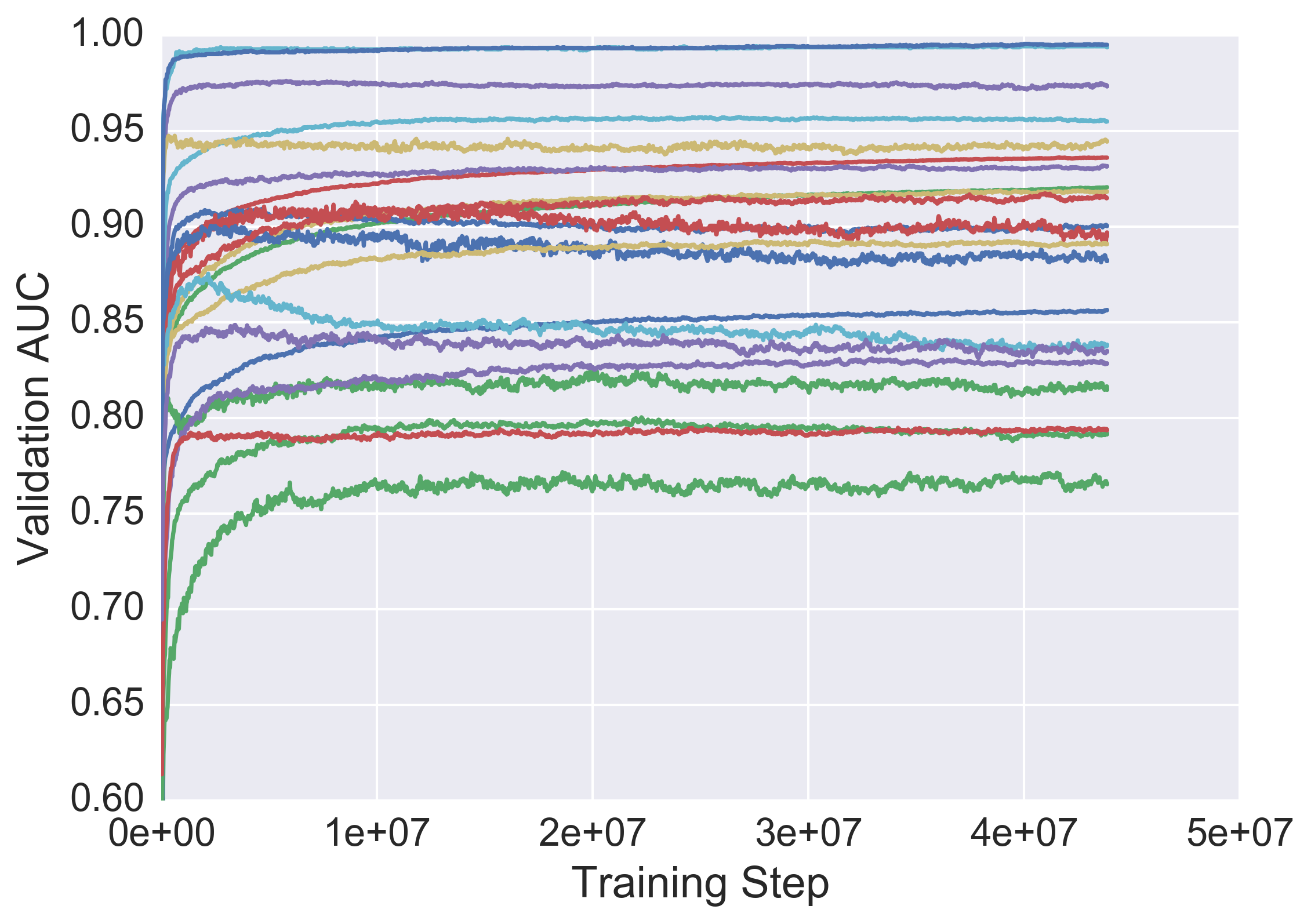}
  \caption{(2000,~1000) W-MTNN+SI}
  \end{subfigure}
  \begin{subfigure}{0.49\linewidth}
  \includegraphics[width=\linewidth]{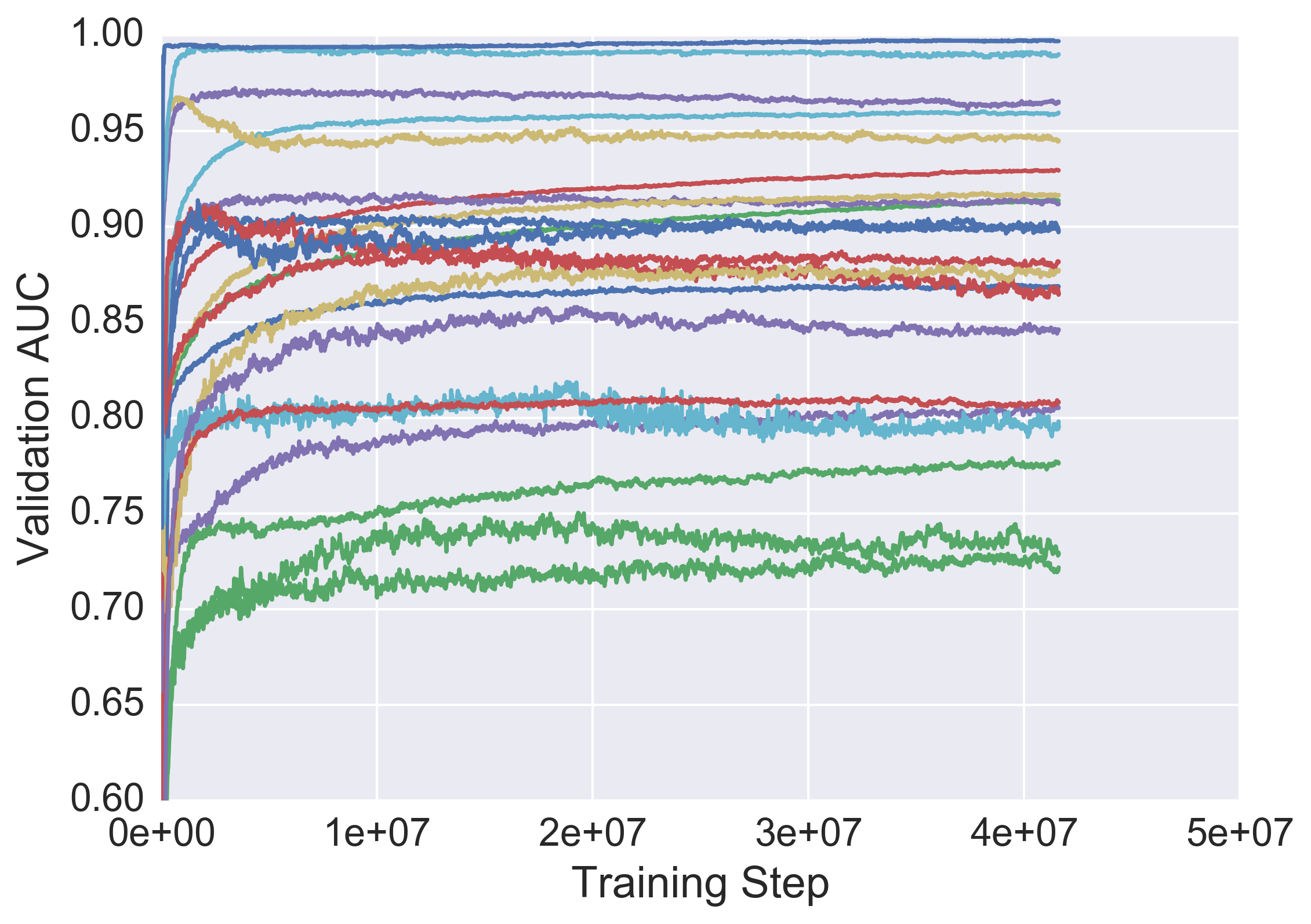}
  \caption{(4000,~2000,~1000,~1000) W-MTNN+SI}
  \end{subfigure}
  \caption{Validation set AUC for each task in task-weighted multitask models
    trained with side information (W-MTNN+SI) as a function of training step.}
  \label{fig:appendix:w_mtnn_training_curves}
\end{figure}

\begin{table*}[htb]
    \caption{Median test set AUC values for MTNN
    models trained with additional side information (SI). We also report
    median $\Delta$AUC values and sign test 95\% confidence intervals for
    comparisons between each model and random forest or logistic regression.
    Bold values indicate confidence intervals that do not include 0.5.}
    \label{appendix:table:public_results}
    \centering
    \rowcolors{2}{lightgray}{}
    \sisetup{detect-weight=true,detect-inline-weight=math}
    \begin{tabular}{ l l S S c S c }
    \toprule
     & & & \multicolumn{2}{c}{MTNN - Random Forest} &
           \multicolumn{2}{c}{MTNN - Logistic Regression} \\
    \cmidrule(lr){4-5} \cmidrule(lr){6-7}
     & Model &
    {\makecell{Median \\ AUC}} &
    {\makecell{Median \\ $\Delta$AUC}} & \makecell{Sign Test \\ 95\% CI} &
    {\makecell{Median \\ $\Delta$AUC}} & \makecell{Sign Test \\ 95\% CI} \\
    \midrule
    \cellcolor{white} & (1000) & 0.773 & 0.030 & (\num{0.47}, \num{0.84}) & 0.003 & (\num{0.31}, \num{0.69}) \\
    \cellcolor{white} & (4000) & 0.824 & \bfseries 0.058 & \bfseries (\num{0.61}, \num{0.93}) & \bfseries 0.035 & \bfseries (\num{0.57}, \num{0.90}) \\
    \cellcolor{white} & (2000, 100) & 0.772 & 0.002 & (\num{0.35}, \num{0.73}) & -0.019 & (\num{0.23}, \num{0.61}) \\
    \cellcolor{white} & (2000, 1000) & 0.809 & \bfseries 0.047 & \bfseries (\num{0.61}, \num{0.93}) & 0.021 & (\num{0.39}, \num{0.77}) \\
    \multirow{-5}{*}{\cellcolor{white} \makecell[l]{U-MTNN\\+SI}} & (4000, 2000, 1000, 1000) & 0.783 & 0.044 & (\num{0.47}, \num{0.84}) & 0.016 & (\num{0.39}, \num{0.77}) \\
    \midrule
    \cellcolor{white} & (1000) & 0.816 & \bfseries 0.061 & \bfseries (\num{0.67}, \num{0.95}) & \bfseries 0.035 & \bfseries (\num{0.61}, \num{0.93}) \\
    \cellcolor{white} & (4000) & 0.808 & \bfseries 0.059 & \bfseries (\num{0.67}, \num{0.95}) & \bfseries 0.032 & \bfseries (\num{0.67}, \num{0.95}) \\
    \cellcolor{white} & (2000, 100) & 0.825 & \bfseries 0.054 & \bfseries (\num{0.67}, \num{0.95}) & \bfseries 0.035 & \bfseries (\num{0.57}, \num{0.90}) \\
    \cellcolor{white} & (2000, 1000) & 0.840 & \bfseries 0.072 & \bfseries (\num{0.78}, \num{0.99}) & \bfseries 0.059 & \bfseries (\num{0.61}, \num{0.93}) \\
    \multirow{-5}{*}{\cellcolor{white} \makecell[l]{W-MTNN\\+SI}} & (4000, 2000, 1000, 1000) & 0.837 & \bfseries 0.074 & \bfseries (\num{0.67}, \num{0.95}) & \bfseries 0.062 & \bfseries (\num{0.57}, \num{0.90}) \\
    \bottomrule
    \end{tabular}
\end{table*}

\begin{table*}[tbp]
    \caption{Comparisons between neural network models trained with additional side
    information (SI). Differences between STNN, U-MTNN+SI, and W-MTNN+SI models
    with the same core architecture are reported as median $\Delta$AUC values
    and sign test 95\% confidence intervals. Bold values indicate confidence
    intervals that do not include 0.5.}
    \label{appendix:table:public_mtnn_vs_stnn}
    \centering
    \rowcolors{2}{lightgray}{}
    \sisetup{detect-weight=true,detect-inline-weight=math}
    \begin{tabular}{ l l S c S c }
    \toprule
     & & \multicolumn{2}{c}{MTNN+SI - STNN} &
         \multicolumn{2}{c}{W-MTNN+SI - U-MTNN+SI} \\
    \cmidrule(lr){3-4} \cmidrule(lr){5-6}
     & Model &
    {\makecell{Median \\ $\Delta$AUC}} & \makecell{Sign Test \\ 95\% CI} &
    {\makecell{Median \\ $\Delta$AUC}} & \makecell{Sign Test \\ 95\% CI} \\
    \midrule
    \cellcolor{white} & (1000) & -0.008 & (\num{0.27}, \num{0.65}) &  &  \\
    \cellcolor{white} & (4000) & 0.008 & (\num{0.43}, \num{0.80}) &  &  \\
    \cellcolor{white} & (2000, 100) & -0.024 & (\num{0.27}, \num{0.65}) &  &  \\
    \cellcolor{white} & (2000, 1000) & 0.018 & (\num{0.39}, \num{0.77}) &  &  \\
    \multirow{-5}{*}{\cellcolor{white} \makecell[l]{U-MTNN\\+SI}} & (4000, 2000, 1000, 1000) & 0.015 & (\num{0.39}, \num{0.77}) &  &  \\
    \midrule
    \cellcolor{white} & (1000) & 0.025 & (\num{0.39}, \num{0.77}) & \bfseries 0.034 & \bfseries (\num{0.78}, \num{0.99}) \\
    \cellcolor{white} & (4000) & \bfseries 0.007 & \bfseries (\num{0.52}, \num{0.87}) & 0.005 & (\num{0.47}, \num{0.84}) \\
    \cellcolor{white} & (2000, 100) & 0.033 & (\num{0.43}, \num{0.80}) & \bfseries 0.048 & \bfseries (\num{0.72}, \num{0.97}) \\
    \cellcolor{white} & (2000, 1000) & \bfseries 0.045 & \bfseries (\num{0.61}, \num{0.93}) & \bfseries 0.020 & \bfseries (\num{0.67}, \num{0.95}) \\
    \multirow{-5}{*}{\cellcolor{white} \makecell[l]{W-MTNN\\+SI}} & (4000, 2000, 1000, 1000) & \bfseries 0.043 & \bfseries (\num{0.57}, \num{0.90}) & \bfseries 0.036 & \bfseries (\num{0.52}, \num{0.87}) \\
    \bottomrule
    \end{tabular}
\end{table*}

\begin{figure}[htbp]
  \centering
  \includegraphics[width=0.5\linewidth]{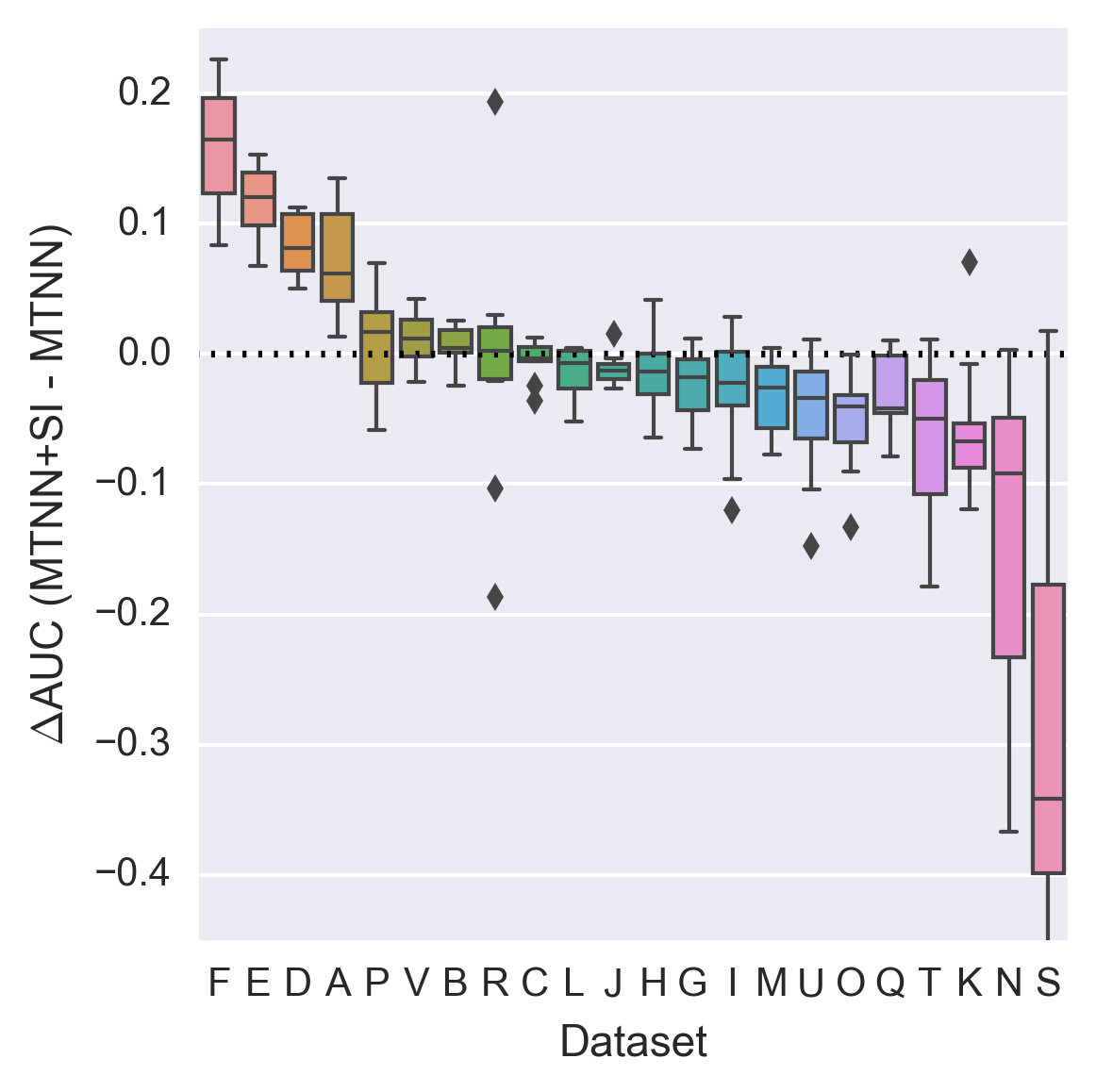}
  \caption{Differences in test set AUC values between MTNN models
  trained with and without side information (SI). Each box plot summarizes 10
  $\Delta$AUC values, one for each combination of model architecture
  (e.g.~(2000,~1000)) and task weighting strategy (U-MTNN or W-MTNN).}
  \label{fig:appendix:mtnn_vs_side}
\end{figure}

\subsubsection{Task relatedness}
\label{sec:appendix:relatedness}

\begin{table}[htb]
    \caption{Datasets used to construct subset multitask models. Each subset
    multitask model used a subset of the datasets from
    \tablename~\ref{table:datasets}. The datasets in each subset are related by
    a similar assay target.}
    \label{appendix:table:subset_datasets}
    \centering
    \rowcolors{1}{}{lightgray}
    \begin{tabular}{ l l }
    \toprule
    Subset & Datasets \\
    \midrule
    Solubility & {B, C} \\
    Metabolism & {D, E, F} \\
    Stability & {N, O, Q} \\
    Clearance & {T, U} \\
    \bottomrule
    \end{tabular}
\end{table}

\begin{table*}[htb]
    \caption{Comparisons between subset and full multitask models. We report the
    median $\Delta$AUC and sign test 95\% confidence interval comparing subset
    and full multitask models with the same core architecture and task weighting
    strategy. Differences were calculated only for the 10 datasets that were used to
    build subset models (see
    \tablename~\ref{appendix:table:subset_datasets}). Bold values indicate
    confidence intervals that do not include 0.5.}
    \label{appendix:table:subset_deltas}
    \centering
    \rowcolors{1}{}{lightgray}
    \sisetup{detect-all=true}
    \begin{tabular}{ l l S c }
    \toprule
     & & \multicolumn{2}{c}{Subset MTNN - Full MTNN} \\
    \cmidrule(lr){3-4}
     & Model &
    {\makecell{Median \\ $\Delta$AUC}} &
    \makecell{Sign Test \\ 95\% CI} \\
    \midrule
    \cellcolor{white} & (1000) & -0.005 & (\num{0.11}, \num{0.60}) \\
    \cellcolor{white} & (4000) & -0.005 & (\num{0.06}, \num{0.51}) \\
    \cellcolor{white} & (2000, 100) & -0.009 & (\num{0.17}, \num{0.69}) \\
    \cellcolor{white} & (2000, 1000) & -0.015 & (\num{0.17}, \num{0.69}) \\
    \multirow{-5}{*}{\cellcolor{white} U-MTNN} & (4000, 2000, 1000, 1000) & -0.016 & (\num{0.17}, \num{0.69}) \\
    \midrule
    \cellcolor{white} & (1000) & -0.005 & (\num{0.06}, \num{0.51}) \\
    \cellcolor{white} & (4000) & -0.006 & (\num{0.11}, \num{0.60}) \\
    \cellcolor{white} & (2000, 100) & -0.008 & (\num{0.17}, \num{0.69}) \\
    \cellcolor{white} & (2000, 1000) & -0.015 & (\num{0.11}, \num{0.60}) \\
    \multirow{-5}{*}{\cellcolor{white} W-MTNN} & (4000, 2000, 1000, 1000) & -0.020 & (\num{0.17}, \num{0.69}) \\
    \bottomrule
    \end{tabular}
\end{table*}

\begin{figure}[htbp]
  \centering
  \includegraphics[width=0.5\linewidth]{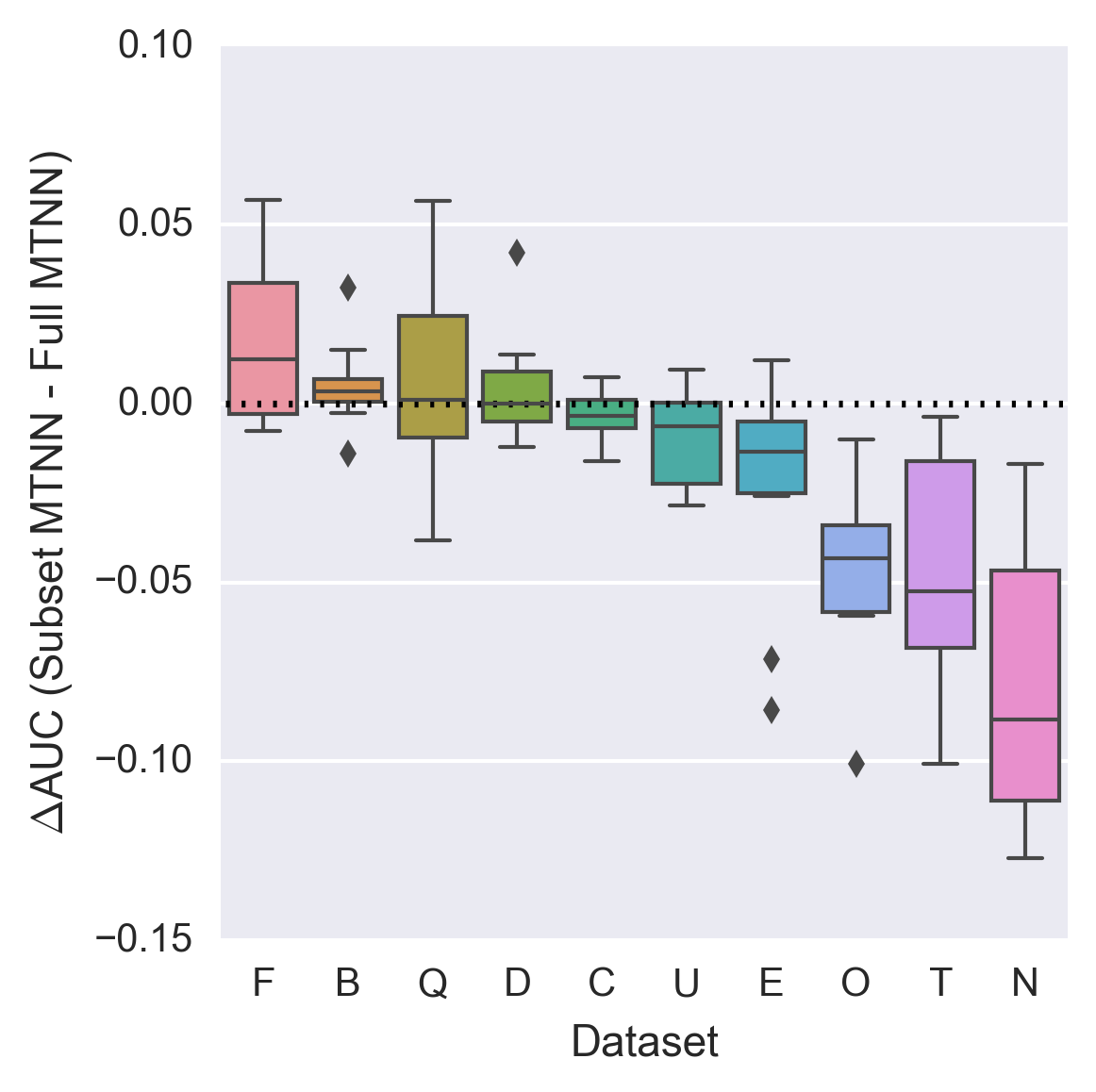}
  \caption{Box plots showing $\Delta$AUC values between subset and full models
  with the same core architecture. Each box plot summarizes 10 $\Delta$AUC
  values, one for each combination of model architecture (e.g.~(2000,~1000)) and
  task weighting strategy (U-MTNN or W-MTNN).}
  \label{fig:appendix:subset_vs_full_subset}
\end{figure}

Multitask learning can only improve performance when tasks are
related~\citep{caruana1997multitask}. Unfortunately, the concept of
\emph{relatedness} is not well defined for chemical datasets. Any measure of
task relatedness must capture the degree to which two datasets follow the
\emph{similar property principle}~\citep{willett2009similarity} for ligand-based
virtual screening: compounds with similar properties should have similar
behavior. Hence, both the features used as input to the model (molecular
descriptors) and the training labels (experimental behavior) must be considered.
Previous attempts to quantify relatedness have focused on shared compounds or
correlations between labels~\citep{erhan2006collaborative,
ramsundar2015massively}. We attempted to construct a new metric for task
relatedness based on correlation between labels for similar compounds in
different tasks (described below), but did not fully explore any relationship
between this measure of task relatedness and multitask improvement in temporal
validation.

We use Tanimoto similarity between compound fingerprints to calculate a measure
of dataset relatedness $R$ under the assumption that similar compounds will have
similar labels in related tasks. Given two datasets, $\alpha$ and $\beta$, we
calculate the Tanimoto similarity ($T_{a,b}$) for each pair of compounds
$(a\in\alpha,b\in\beta)$. For pairs above a similarity threshold $\tau$, we
count the number of pairs with the same ($S$) or different ($D$) labels. Since
label meanings are somewhat arbitrary---related tasks may have correlated or
anticorrelated labels---we then take the maximum of $S$ and $D$ and normalize it
by the number of similar pairs (non-similar pairs provide no information about
task relatedness). Formally:
\begin{align}
S(\alpha, \beta) &= \sum_{a \in \alpha, b \in \beta} \mathbbm{1} \left(T_{a,b} \ge \tau \right) \mathbbm{1} \left(y_a = y_b\right) \\
D(\alpha, \beta) &= \sum_{a \in \alpha, b \in \beta} \mathbbm{1} \left(T_{a,b} \ge \tau \right) \mathbbm{1} \left(y_a \neq y_b\right) \\
R(\alpha, \beta) &= \frac{\max\left\{S(\alpha, \beta), D(\alpha, \beta)\right\}}{S(\alpha, \beta) + D(\alpha, \beta)}
\end{align}
Note that this metric is symmetric and takes into account both the number of
similar compounds and the correlation of their labels. It ranges from 0.5 to
1.0, with unity indicating identical datasets. We used $\tau=0.5$, in accordance
with the similarity/dissimiliarity threshold for ECFP4 estimated
by~\citet{franco2014use}. This measure of relatedness has some
shortcomings, including a tendency to give nonintuitive results when datasets
have very few similar pairs.

One potential problem with attempts to correlate any measure of task relatedness
with multitask improvement is that, in general, multitask improvements are
relatively small ($<10\%$) and dataset-dependent, which makes it somewhat
dangerous to look for correlations at all.

\subsection{Information leakage in multitask networks}

\begin{table*}[htb]
    \caption{Median test set AUC values for models using non-leaky temporal
    validation. We also report median $\Delta$AUC values and sign test 95\%
    confidence intervals for comparisons between each model and random forest or
    logistic regression. Bold values indicate confidence intervals that do not
    include 0.5.}
    \label{appendix:table:focused_results}
    \centering
    \rowcolors{2}{lightgray}{}
    \sisetup{detect-weight=true,detect-inline-weight=math}
    \begin{tabular}{ l l S S c S c }
    \toprule
     & & & \multicolumn{2}{c}{Model - Random Forest} &
           \multicolumn{2}{c}{Model - Logistic Regression} \\
    \cmidrule(lr){4-5} \cmidrule(lr){6-7}
     & Model &
    {\makecell{Median \\ AUC}} &
    {\makecell{Median \\ $\Delta$AUC}} & \makecell{Sign Test \\ 95\% CI} &
    {\makecell{Median \\ $\Delta$AUC}} & \makecell{Sign Test \\ 95\% CI} \\
    \midrule
    \cellcolor{white} & (1000) & 0.793 & \bfseries 0.056 & \bfseries (\num{0.67}, \num{0.95}) & \bfseries 0.036 & \bfseries (\num{0.57}, \num{0.90}) \\
    \cellcolor{white} & (4000) & 0.789 & \bfseries 0.055 & \bfseries (\num{0.67}, \num{0.95}) & \bfseries 0.029 & \bfseries (\num{0.57}, \num{0.90}) \\
    \cellcolor{white} & (2000, 100) & 0.759 & \bfseries 0.040 & \bfseries (\num{0.57}, \num{0.90}) & 0.014 & (\num{0.47}, \num{0.84}) \\
    \cellcolor{white} & (2000, 1000) & 0.778 & \bfseries 0.049 & \bfseries (\num{0.65}, \num{0.95}) & \bfseries 0.037 & \bfseries (\num{0.61}, \num{0.93}) \\
    \multirow{-5}{*}{\cellcolor{white} U-MTNN} & (4000, 2000, 1000, 1000) & 0.784 & \bfseries 0.055 & \bfseries (\num{0.72}, \num{0.97}) & 0.033 & (\num{0.47}, \num{0.84}) \\
    \midrule
    \cellcolor{white} & (1000) & 0.793 & \bfseries 0.059 & \bfseries (\num{0.61}, \num{0.93}) & \bfseries 0.039 & \bfseries (\num{0.67}, \num{0.95}) \\
    \cellcolor{white} & (4000) & 0.783 & \bfseries 0.058 & \bfseries (\num{0.61}, \num{0.93}) & \bfseries 0.025 & \bfseries (\num{0.67}, \num{0.95}) \\
    \cellcolor{white} & (2000, 100) & 0.761 & \bfseries 0.034 & \bfseries (\num{0.61}, \num{0.93}) & 0.024 & (\num{0.43}, \num{0.80}) \\
    \cellcolor{white} & (2000, 1000) & 0.782 & \bfseries 0.055 & \bfseries (\num{0.67}, \num{0.95}) & \bfseries 0.040 & \bfseries (\num{0.61}, \num{0.93}) \\
    \multirow{-5}{*}{\cellcolor{white} W-MTNN} & (4000, 2000, 1000, 1000) & 0.785 & \bfseries 0.051 & \bfseries (\num{0.61}, \num{0.93}) & 0.039 & (\num{0.47}, \num{0.84}) \\
    \bottomrule
    \end{tabular}
\end{table*}

\begin{table*}[tbp]
    \caption{Comparisons between neural network models using non-leaky temporal
    validation. Differences between STNN, U-MTNN, and W-MTNN models with the same
    core architecture are reported as median $\Delta$AUC values and sign test
    95\% confidence intervals. Bold values indicate confidence intervals that do
    not include 0.5.}
    \label{table:appendix:focused_mtnn_vs_stnn}
    \centering
    \rowcolors{2}{lightgray}{}
    \sisetup{detect-weight=true,detect-inline-weight=math}
    \begin{tabular}{ l l S c S c }
    \toprule
     & & \multicolumn{2}{c}{MTNN - STNN} &
         \multicolumn{2}{c}{W-MTNN - U-MTNN} \\
    \cmidrule(lr){3-4} \cmidrule(lr){5-6}
     & Model &
    {\makecell{Median \\ $\Delta$AUC}} & \makecell{Sign Test \\ 95\% CI} &
    {\makecell{Median \\ $\Delta$AUC}} & \makecell{Sign Test \\ 95\% CI} \\
    \midrule
    \cellcolor{white} & (1000) & \bfseries 0.009 & \bfseries (\num{0.57}, \num{0.90}) &  &  \\
    \cellcolor{white} & (4000) & 0.004 & (\num{0.43}, \num{0.80}) &  &  \\
    \cellcolor{white} & (2000, 100) & 0.012 & (\num{0.47}, \num{0.84}) &  &  \\
    \cellcolor{white} & (2000, 1000) & \bfseries 0.027 & \bfseries (\num{0.57}, \num{0.90}) &  &  \\
    \multirow{-5}{*}{\cellcolor{white} U-MTNN} & (4000, 2000, 1000, 1000) & 0.029 & (\num{0.43}, \num{0.80}) &  &  \\
    \midrule
    \cellcolor{white} & (1000) & \bfseries 0.017 & \bfseries (\num{0.67}, \num{0.95}) & -0.000 & (\num{0.31}, \num{0.69}) \\
    \cellcolor{white} & (4000) & \bfseries 0.008 & \bfseries (\num{0.52}, \num{0.87}) & 0.001 & (\num{0.35}, \num{0.73}) \\
    \cellcolor{white} & (2000, 100) & 0.015 & (\num{0.39}, \num{0.77}) & 0.005 & (\num{0.43}, \num{0.80}) \\
    \cellcolor{white} & (2000, 1000) & \bfseries 0.026 & \bfseries (\num{0.52}, \num{0.87}) & -0.001 & (\num{0.27}, \num{0.65}) \\
    \multirow{-5}{*}{\cellcolor{white} W-MTNN} & (4000, 2000, 1000, 1000) & 0.032 & (\num{0.47}, \num{0.84}) & \bfseries 0.011 & \bfseries (\num{0.52}, \num{0.87}) \\
    \bottomrule
    \end{tabular}
\end{table*}

\begin{table*}[htb]
    \caption{Pairwise comparisons between neural network model architectures
    using non-leaky temporal validation. For each pair of models within a model
    class (e.g.~STNN), we report the median $\Delta$AUC and sign test
    95\% confidence interval. Bold values indicate confidence intervals that do
    not include 0.5.}
    \label{appendix:table:architecture_comparison_temporal}
    \centering
    \rowcolors{2}{}{lightgray}
    \sisetup{detect-weight=true,detect-inline-weight=math}
    \begin{tabular}{ l l l S c }
    \toprule
     & & & \multicolumn{2}{c}{Model B - Model A} \\
    \cmidrule(lr){4-5}
     & Model A & Model B &
    {\makecell{Median \\ $\Delta$AUC}} &
    \makecell{Sign Test \\ 95\% CI} \\
    \midrule
    \cellcolor{white} & (1000) & (4000) & \bfseries 0.006 & \bfseries (\num{0.61}, \num{0.93}) \\
    \cellcolor{white} & (1000) & (2000, 100) & -0.004 & (\num{0.23}, \num{0.61}) \\
    \cellcolor{white} & (1000) & (2000, 1000) & -0.004 & (\num{0.20}, \num{0.57}) \\
    \cellcolor{white} & (1000) & (4000, 2000, 1000, 1000) & -0.004 & (\num{0.27}, \num{0.65}) \\
    \cellcolor{white} & (4000) & (2000, 100) & \bfseries \color{red} -0.006 & \bfseries \color{red} (\num{0.13}, \num{0.48}) \\
    \cellcolor{white} & (4000) & (2000, 1000) & -0.006 & (\num{0.20}, \num{0.57}) \\
    \cellcolor{white} & (4000) & (4000, 2000, 1000, 1000) & \bfseries \color{red} -0.011 & \bfseries \color{red} (\num{0.13}, \num{0.48}) \\
    \cellcolor{white} & (2000, 100) & (2000, 1000) & 0.002 & (\num{0.35}, \num{0.73}) \\
    \cellcolor{white} & (2000, 100) & (4000, 2000, 1000, 1000) & -0.001 & (\num{0.28}, \num{0.68}) \\
    \multirow{-10}{*}{\cellcolor{white} STNN} & (2000, 1000) & (4000, 2000, 1000, 1000) & -0.006 & (\num{0.16}, \num{0.53}) \\
    \midrule
    \cellcolor{white} & (1000) & (4000) & -0.001 & (\num{0.27}, \num{0.65}) \\
    \cellcolor{white} & (1000) & (2000, 100) & -0.012 & (\num{0.23}, \num{0.61}) \\
    \cellcolor{white} & (1000) & (2000, 1000) & 0.004 & (\num{0.43}, \num{0.80}) \\
    \cellcolor{white} & (1000) & (4000, 2000, 1000, 1000) & -0.001 & (\num{0.27}, \num{0.65}) \\
    \cellcolor{white} & (4000) & (2000, 100) & -0.008 & (\num{0.23}, \num{0.61}) \\
    \cellcolor{white} & (4000) & (2000, 1000) & 0.008 & (\num{0.43}, \num{0.80}) \\
    \cellcolor{white} & (4000) & (4000, 2000, 1000, 1000) & -0.002 & (\num{0.27}, \num{0.65}) \\
    \cellcolor{white} & (2000, 100) & (2000, 1000) & 0.006 & (\num{0.43}, \num{0.80}) \\
    \cellcolor{white} & (2000, 100) & (4000, 2000, 1000, 1000) & \bfseries 0.007 & \bfseries (\num{0.57}, \num{0.90}) \\
    \multirow{-10}{*}{\cellcolor{white} U-MTNN} & (2000, 1000) & (4000, 2000, 1000, 1000) & -0.001 & (\num{0.27}, \num{0.65}) \\
    \midrule
    \cellcolor{white} & (1000) & (4000) & -0.003 & (\num{0.23}, \num{0.61}) \\
    \cellcolor{white} & (1000) & (2000, 100) & -0.006 & (\num{0.16}, \num{0.53}) \\
    \cellcolor{white} & (1000) & (2000, 1000) & 0.003 & (\num{0.39}, \num{0.77}) \\
    \cellcolor{white} & (1000) & (4000, 2000, 1000, 1000) & -0.005 & (\num{0.20}, \num{0.57}) \\
    \cellcolor{white} & (4000) & (2000, 100) & 0.000 & (\num{0.31}, \num{0.69}) \\
    \cellcolor{white} & (4000) & (2000, 1000) & 0.009 & (\num{0.39}, \num{0.77}) \\
    \cellcolor{white} & (4000) & (4000, 2000, 1000, 1000) & 0.002 & (\num{0.31}, \num{0.69}) \\
    \cellcolor{white} & (2000, 100) & (2000, 1000) & \bfseries 0.015 & \bfseries (\num{0.52}, \num{0.87}) \\
    \cellcolor{white} & (2000, 100) & (4000, 2000, 1000, 1000) & 0.016 & (\num{0.47}, \num{0.84}) \\
    \multirow{-10}{*}{\cellcolor{white} W-MTNN} & (2000, 1000) & (4000, 2000, 1000, 1000) & -0.007 & (\num{0.23}, \num{0.61}) \\
    \bottomrule
    \end{tabular}
\end{table*}

\begin{figure}[tb]
  \centering
  \includegraphics[width=0.5\linewidth]{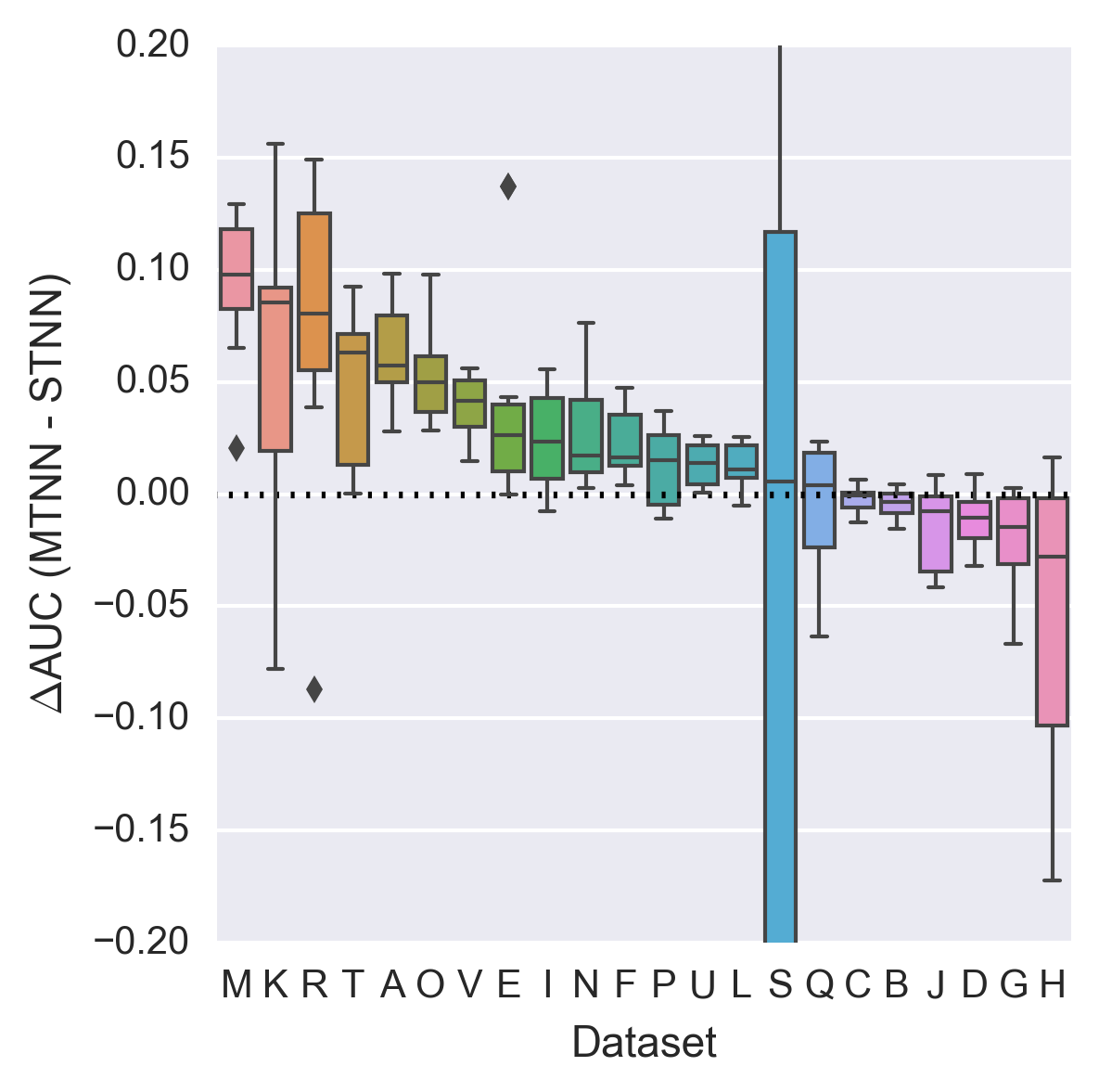}
  \caption{Box plots showing $\Delta$AUC values between non-leaky MTNN models
  and STNN models with the same core architecture. Each box
  plot summarizes 10 $\Delta$AUC values, one for each combination of model
  architecture (e.g.~(2000,~1000)) and task weighting strategy (U-MTNN or
  W-MTNN).}
  \label{fig:multitask_effect_box_temporal}
\end{figure}

\begin{figure}[htbp]
  \centering
  \includegraphics[width=0.5\linewidth]{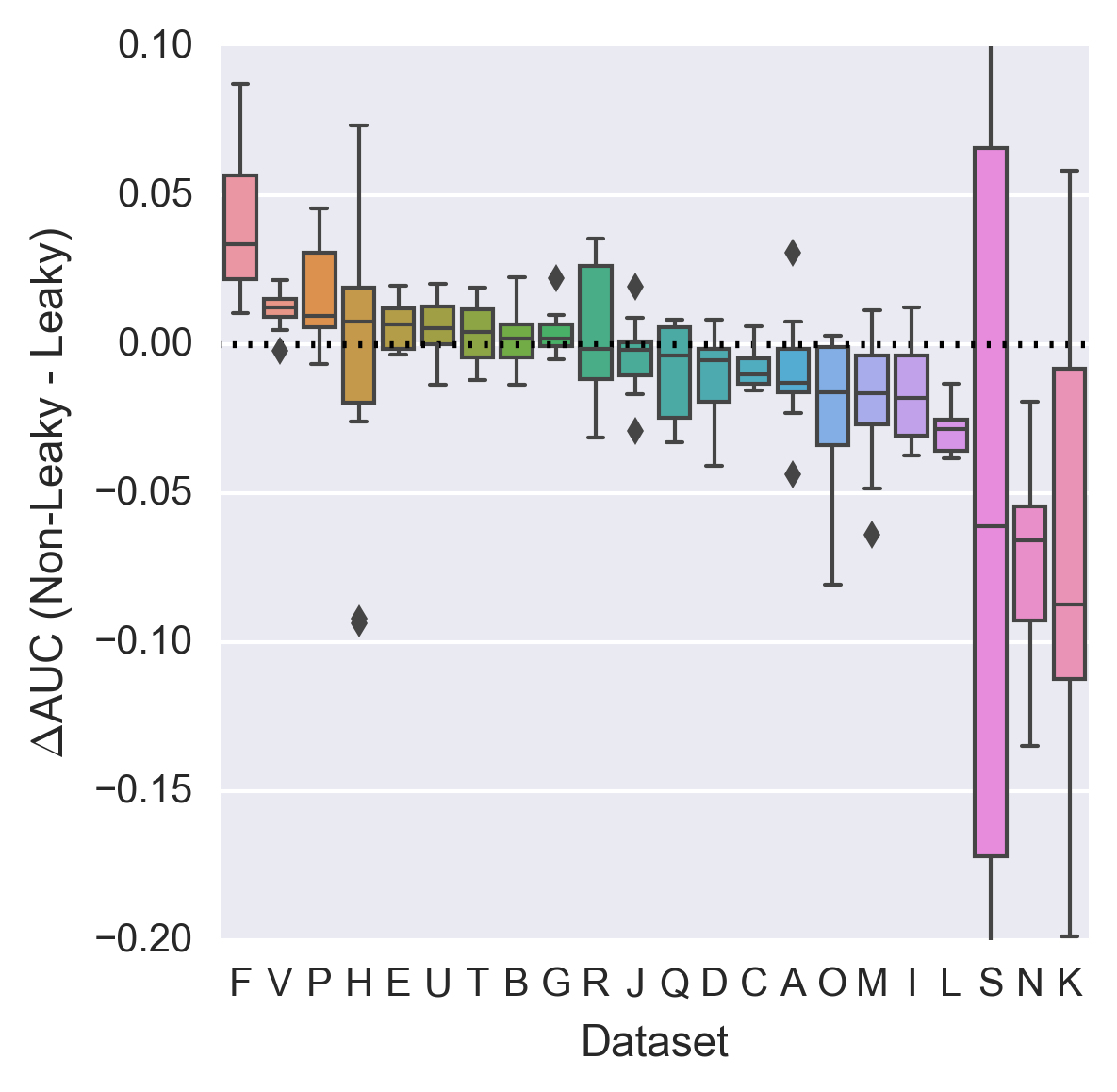}
  \caption{Box plots showing $\Delta$AUC values between non-leaky and leaky
  models with the same core architecture. Each box plot summarizes 10
  $\Delta$AUC values, one for each combination of model architecture
  (e.g.~(2000,~1000)) and task weighting strategy (U-MTNN or W-MTNN).}
  \label{fig:appendix:non_leaky_vs_leaky}
\end{figure}

\subsection{Random cross-validation}
\label{sec:random_cv}

\begin{table*}[htb]
    \caption{Median 5-fold mean test set AUC values for models using random
    cross-validation. We also report median $\Delta$AUC values and sign test
    95\% confidence intervals for comparisons between each model and random
    forest or logistic regression. Bold values indicate confidence intervals
    that do not include 0.5.}
    \label{appendix:table:random_results}
    \centering
    \rowcolors{2}{lightgray}{}
    \sisetup{detect-weight=true,detect-inline-weight=math}
    \begin{tabular}{ l l S S c S c }
    \toprule
     & & & \multicolumn{2}{c}{Model - Random Forest} &
           \multicolumn{2}{c}{Model - Logistic Regression} \\
    \cmidrule(lr){4-5} \cmidrule(lr){6-7}
     & Model &
    {\makecell{Median \\ AUC}} &
    {\makecell{Median \\ $\Delta$AUC}} & \makecell{Sign Test \\ 95\% CI} &
    {\makecell{Median \\ $\Delta$AUC}} & \makecell{Sign Test \\ 95\% CI} \\
    \midrule
    \cellcolor{white} & Random Forest & 0.916 &  &  & \bfseries 0.022 & \bfseries (\num{0.67}, \num{0.95}) \\
    \cellcolor{white} & Logistic Regression & 0.896 & \bfseries \color{red} -0.022 & \bfseries \color{red} (\num{0.05}, \num{0.33}) &  &  \\
    \midrule
    \cellcolor{white} & (1000) & 0.923 & 0.005 & (\num{0.43}, \num{0.80}) & \bfseries 0.021 & \bfseries (\num{0.57}, \num{0.90}) \\
    \cellcolor{white} & (4000) & 0.926 & \bfseries 0.008 & \bfseries (\num{0.61}, \num{0.93}) & \bfseries 0.025 & \bfseries (\num{0.67}, \num{0.95}) \\
    \cellcolor{white} & (2000, 100) & 0.918 & 0.005 & (\num{0.47}, \num{0.84}) & \bfseries 0.024 & \bfseries (\num{0.61}, \num{0.93}) \\
    \cellcolor{white} & (2000, 1000) & 0.915 & 0.002 & (\num{0.35}, \num{0.73}) & \bfseries 0.018 & \bfseries (\num{0.57}, \num{0.90}) \\
    \multirow{-5}{*}{\cellcolor{white} STNN} & (4000, 2000, 1000, 1000) & 0.908 & -0.005 & (\num{0.27}, \num{0.65}) & 0.009 & (\num{0.35}, \num{0.73}) \\
    \midrule
    \cellcolor{white} & (1000) & 0.930 & \bfseries 0.020 & \bfseries (\num{0.85}, \num{1.00}) & \bfseries 0.040 & \bfseries (\num{0.85}, \num{1.00}) \\
    \cellcolor{white} & (4000) & 0.931 & \bfseries 0.020 & \bfseries (\num{0.85}, \num{1.00}) & \bfseries 0.043 & \bfseries (\num{0.85}, \num{1.00}) \\
    \cellcolor{white} & (2000, 100) & 0.930 & \bfseries 0.019 & \bfseries (\num{0.72}, \num{0.97}) & \bfseries 0.034 & \bfseries (\num{0.85}, \num{1.00}) \\
    \cellcolor{white} & (2000, 1000) & 0.939 & \bfseries 0.027 & \bfseries (\num{0.85}, \num{1.00}) & \bfseries 0.051 & \bfseries (\num{0.85}, \num{1.00}) \\
    \multirow{-5}{*}{\cellcolor{white} U-MTNN} & (4000, 2000, 1000, 1000) & 0.937 & \bfseries 0.024 & \bfseries (\num{0.85}, \num{1.00}) & \bfseries 0.042 & \bfseries (\num{0.85}, \num{1.00}) \\
    \midrule
    \cellcolor{white} & (1000) & 0.934 & \bfseries 0.017 & \bfseries (\num{0.85}, \num{1.00}) & \bfseries 0.041 & \bfseries (\num{0.85}, \num{1.00}) \\
    \cellcolor{white} & (4000) & 0.933 & \bfseries 0.019 & \bfseries (\num{0.85}, \num{1.00}) & \bfseries 0.043 & \bfseries (\num{0.85}, \num{1.00}) \\
    \cellcolor{white} & (2000, 100) & 0.931 & \bfseries 0.019 & \bfseries (\num{0.67}, \num{0.95}) & \bfseries 0.042 & \bfseries (\num{0.78}, \num{0.99}) \\
    \cellcolor{white} & (2000, 1000) & 0.936 & \bfseries 0.026 & \bfseries (\num{0.85}, \num{1.00}) & \bfseries 0.050 & \bfseries (\num{0.78}, \num{0.99}) \\
    \multirow{-5}{*}{\cellcolor{white} W-MTNN} & (4000, 2000, 1000, 1000) & 0.937 & \bfseries 0.023 & \bfseries (\num{0.85}, \num{1.00}) & \bfseries 0.046 & \bfseries (\num{0.78}, \num{0.99}) \\
    \bottomrule
    \end{tabular}
\end{table*}

\begin{table*}[htbp]
    \caption{Comparisons between neural network models using random cross-validation.
    Differences between STNN, U-MTNN, and W-MTNN models with the same core
    architecture are reported as median $\Delta$AUC values and sign test 95\%
    confidence intervals. Bold values indicate confidence intervals that do not
    include 0.5.}
    \label{appendix:table:random_mtnn_vs_stnn}
    \centering
    \rowcolors{2}{lightgray}{}
    \sisetup{detect-all=true}
    \begin{tabular}{ l l S c S c }
    \toprule
     & & \multicolumn{2}{c}{MTNN - STNN} &
         \multicolumn{2}{c}{W-MTNN - U-MTNN} \\
    \cmidrule(lr){3-4} \cmidrule(lr){5-6}
     & Model &
    {\makecell{Median \\ $\Delta$AUC}} & \makecell{Sign Test \\ 95\% CI} &
    {\makecell{Median \\ $\Delta$AUC}} & \makecell{Sign Test \\ 95\% CI} \\
    \midrule
    \cellcolor{white} & (1000) & \bfseries 0.012 & \bfseries (\num{0.85}, \num{1.00}) &  &  \\
    \cellcolor{white} & (4000) & \bfseries 0.011 & \bfseries (\num{0.78}, \num{0.99}) &  &  \\
    \cellcolor{white} & (2000, 100) & \bfseries 0.016 & \bfseries (\num{0.57}, \num{0.90}) &  &  \\
    \cellcolor{white} & (2000, 1000) & \bfseries 0.024 & \bfseries (\num{0.85}, \num{1.00}) &  &  \\
    \multirow{-5}{*}{\cellcolor{white} U-MTNN} & (4000, 2000, 1000, 1000) & \bfseries 0.033 & \bfseries (\num{0.85}, \num{1.00}) &  &  \\
    \midrule
    \cellcolor{white} & (1000) & \bfseries 0.012 & \bfseries (\num{0.67}, \num{0.95}) & 0.001 & (\num{0.39}, \num{0.77}) \\
    \cellcolor{white} & (4000) & \bfseries 0.010 & \bfseries (\num{0.85}, \num{1.00}) & 0.000 & (\num{0.39}, \num{0.77}) \\
    \cellcolor{white} & (2000, 100) & \bfseries 0.015 & \bfseries (\num{0.57}, \num{0.90}) & 0.003 & (\num{0.35}, \num{0.73}) \\
    \cellcolor{white} & (2000, 1000) & \bfseries 0.025 & \bfseries (\num{0.85}, \num{1.00}) & -0.001 & (\num{0.20}, \num{0.57}) \\
    \multirow{-5}{*}{\cellcolor{white} W-MTNN} & (4000, 2000, 1000, 1000) & \bfseries 0.033 & \bfseries (\num{0.72}, \num{0.97}) & 0.000 & (\num{0.39}, \num{0.77}) \\
    \bottomrule
    \end{tabular}
\end{table*}

\begin{table*}[htb]
    \caption{Pairwise comparisons between neural network model architectures
    using random cross-validation. For each pair of models within a model class
    (e.g.~STNN), we report the median $\Delta$AUC and sign test 95\%
    confidence interval. Bold values indicate confidence intervals that do not
    include 0.5.}
    \label{appendix:table:random_architecture_comparison}
    \centering
    \rowcolors{1}{}{lightgray}
    \sisetup{detect-weight=true,detect-inline-weight=math}
    \begin{tabular}{ l l l S c }
    \toprule
     & & & \multicolumn{2}{c}{Model B - Model A} \\
    \cmidrule(lr){4-5}
     & Model A & Model B &
    {\makecell{Median \\ $\Delta$AUC}} &
    \makecell{Sign Test \\ 95\% CI} \\
    \midrule
    \cellcolor{white} & (1000) & (4000) & \bfseries 0.003 & \bfseries (\num{0.67}, \num{0.95}) \\
    \cellcolor{white} & (1000) & (2000, 100) & 0.001 & (\num{0.39}, \num{0.77}) \\
    \cellcolor{white} & (1000) & (2000, 1000) & \bfseries \color{red} -0.001 & \bfseries \color{red} (\num{0.13}, \num{0.48}) \\
    \cellcolor{white} & (1000) & (4000, 2000, 1000, 1000) & \bfseries \color{red} -0.007 & \bfseries \color{red} (\num{0.10}, \num{0.43}) \\
    \cellcolor{white} & (4000) & (2000, 100) & \bfseries \color{red} -0.002 & \bfseries \color{red} (\num{0.13}, \num{0.48}) \\
    \cellcolor{white} & (4000) & (2000, 1000) & \bfseries \color{red} -0.005 & \bfseries \color{red} (\num{0.01}, \num{0.22}) \\
    \cellcolor{white} & (4000) & (4000, 2000, 1000, 1000) & \bfseries \color{red} -0.010 & \bfseries \color{red} (\num{0.03}, \num{0.28}) \\
    \cellcolor{white} & (2000, 100) & (2000, 1000) & \bfseries \color{red} -0.003 & \bfseries \color{red} (\num{0.00}, \num{0.15}) \\
    \cellcolor{white} & (2000, 100) & (4000, 2000, 1000, 1000) & \bfseries \color{red} -0.008 & \bfseries \color{red} (\num{0.03}, \num{0.28}) \\
    \multirow{-10}{*}{\cellcolor{white} STNN} & (2000, 1000) & (4000, 2000, 1000, 1000) & -0.003 & (\num{0.16}, \num{0.53}) \\
    \midrule
    \cellcolor{white} & (1000) & (4000) & 0.001 & (\num{0.43}, \num{0.80}) \\
    \cellcolor{white} & (1000) & (2000, 100) & -0.003 & (\num{0.20}, \num{0.57}) \\
    \cellcolor{white} & (1000) & (2000, 1000) & \bfseries 0.009 & \bfseries (\num{0.78}, \num{0.99}) \\
    \cellcolor{white} & (1000) & (4000, 2000, 1000, 1000) & 0.005 & (\num{0.47}, \num{0.84}) \\
    \cellcolor{white} & (4000) & (2000, 100) & -0.001 & (\num{0.27}, \num{0.65}) \\
    \cellcolor{white} & (4000) & (2000, 1000) & \bfseries 0.008 & \bfseries (\num{0.85}, \num{1.00}) \\
    \cellcolor{white} & (4000) & (4000, 2000, 1000, 1000) & \bfseries 0.003 & \bfseries (\num{0.57}, \num{0.90}) \\
    \cellcolor{white} & (2000, 100) & (2000, 1000) & \bfseries 0.010 & \bfseries (\num{0.85}, \num{1.00}) \\
    \cellcolor{white} & (2000, 100) & (4000, 2000, 1000, 1000) & \bfseries 0.005 & \bfseries (\num{0.67}, \num{0.95}) \\
    \multirow{-10}{*}{\cellcolor{white} U-MTNN} & (2000, 1000) & (4000, 2000, 1000, 1000) & \bfseries \color{red} -0.003 & \bfseries \color{red} (\num{0.10}, \num{0.43}) \\
    \midrule
    \cellcolor{white} & (1000) & (4000) & 0.000 & (\num{0.35}, \num{0.73}) \\
    \cellcolor{white} & (1000) & (2000, 100) & 0.001 & (\num{0.35}, \num{0.73}) \\
    \cellcolor{white} & (1000) & (2000, 1000) & \bfseries 0.009 & \bfseries (\num{0.67}, \num{0.95}) \\
    \cellcolor{white} & (1000) & (4000, 2000, 1000, 1000) & \bfseries 0.006 & \bfseries (\num{0.52}, \num{0.87}) \\
    \cellcolor{white} & (4000) & (2000, 100) & -0.002 & (\num{0.23}, \num{0.61}) \\
    \cellcolor{white} & (4000) & (2000, 1000) & \bfseries 0.007 & \bfseries (\num{0.57}, \num{0.90}) \\
    \cellcolor{white} & (4000) & (4000, 2000, 1000, 1000) & 0.003 & (\num{0.35}, \num{0.73}) \\
    \cellcolor{white} & (2000, 100) & (2000, 1000) & \bfseries 0.006 & \bfseries (\num{0.57}, \num{0.90}) \\
    \cellcolor{white} & (2000, 100) & (4000, 2000, 1000, 1000) & \bfseries 0.004 & \bfseries (\num{0.67}, \num{0.95}) \\
    \multirow{-10}{*}{\cellcolor{white} W-MTNN} & (2000, 1000) & (4000, 2000, 1000, 1000) & -0.002 & (\num{0.20}, \num{0.57}) \\
    \bottomrule
    \end{tabular}
\end{table*}

\begin{figure}[tb]
  \centering
  \includegraphics[width=0.5\linewidth]{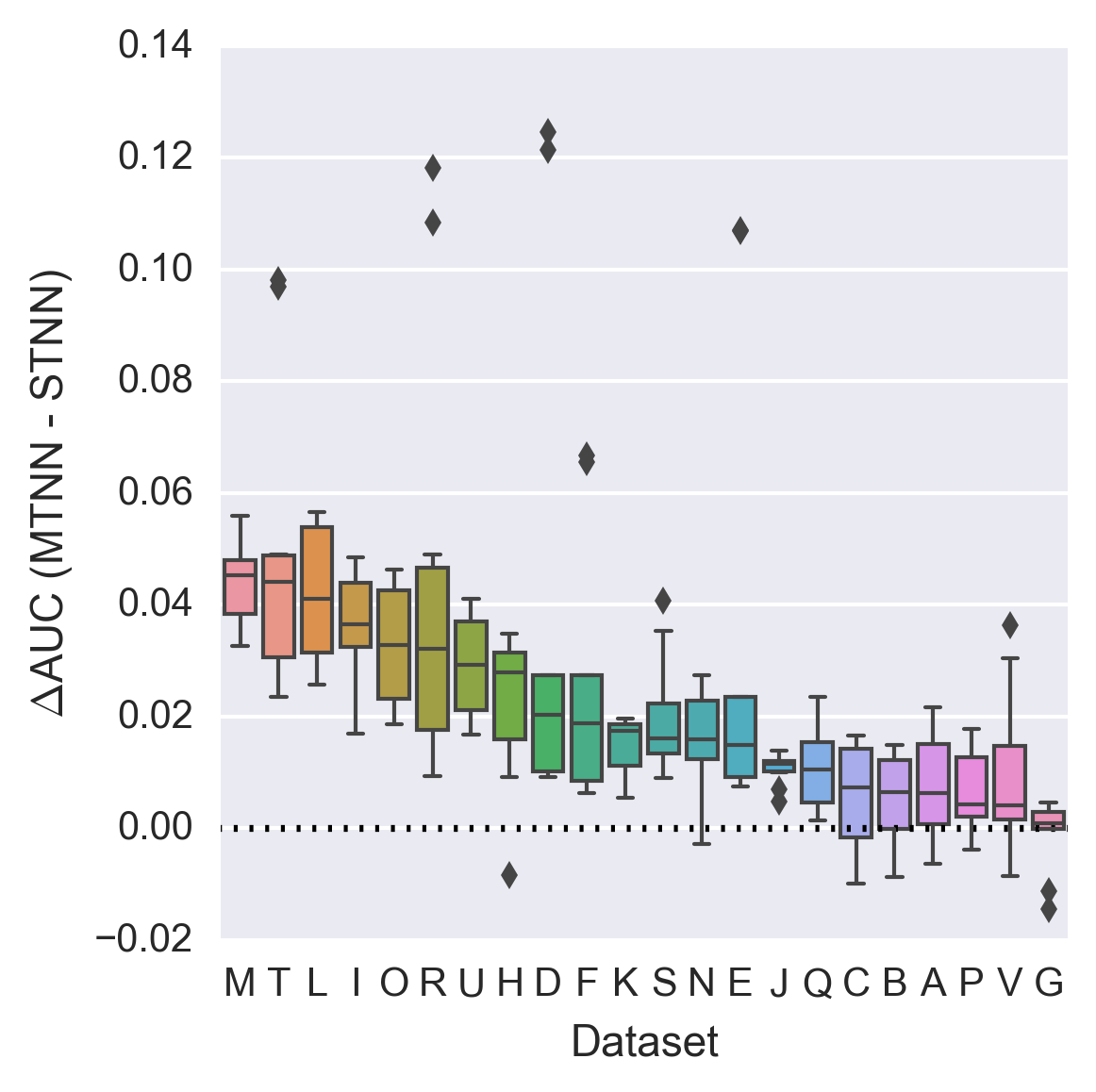}
  \caption{Box plots showing $\Delta$AUC values between MTNN
  and STNN models with the same core architecture for models trained
  using random cross-validation. Each box plot summarizes 10 $\Delta$AUC values,
  one for each combination of model architecture (e.g.~(2000,~1000)) and task
  weighting strategy (U-MTNN or W-MTNN).}
  \label{fig:multitask_effect_box_random}
\end{figure}

\subsection{Random cross-validation: Alternative evaluation strategy}
\label{sec:appendix:random_cv_alternative}

Because we did not hold out a validation set for checkpoint selection when
training random cross-validation models, the values reported in
Section~\ref{sec:random_cv} were calculated from the final training checkpoint of each
per-fold model. In this section, we report values generated using the per-fold
checkpoints closest to a ``target step'' that maximized the 5-fold mean AUC for
each task. Note that target steps were chosen for each task independently.

\begin{table*}[htb]
    \caption{Median 5-fold mean test set AUC values for models using random
    cross-validation with target step evaluation. We also report median
    $\Delta$AUC values and sign test 95\% confidence intervals for comparisons
    between each model and random forest or logistic regression. Bold values
    indicate confidence intervals that do not include 0.5.}
    \label{appendix:table:random_results_alternative}
    \centering
    \rowcolors{2}{lightgray}{}
    \sisetup{detect-weight=true,detect-inline-weight=math}
    \begin{tabular}{ l l S S c S c }
    \toprule
     & & & \multicolumn{2}{c}{Model - Random Forest} &
           \multicolumn{2}{c}{Model - Logistic Regression} \\
    \cmidrule(lr){4-5} \cmidrule(lr){6-7}
     & Model &
    {\makecell{Median \\ AUC}} &
    {\makecell{Median \\ $\Delta$AUC}} & \makecell{Sign Test \\ 95\% CI} &
    {\makecell{Median \\ $\Delta$AUC}} & \makecell{Sign Test \\ 95\% CI} \\
    \midrule
    \cellcolor{white} & Random Forest & 0.916 &  &  & \bfseries 0.022 & \bfseries (\num{0.67}, \num{0.95}) \\
    \cellcolor{white} & Logistic Regression & 0.896 & \bfseries \color{red} -0.022 & \bfseries \color{red} (\num{0.05}, \num{0.33}) &  &  \\
    \midrule
    \cellcolor{white} & (1000) & 0.928 & \bfseries 0.009 & \bfseries (\num{0.72}, \num{0.97}) & \bfseries 0.026 & \bfseries (\num{0.72}, \num{0.97}) \\
    \cellcolor{white} & (4000) & 0.931 & \bfseries 0.010 & \bfseries (\num{0.72}, \num{0.97}) & \bfseries 0.028 & \bfseries (\num{0.78}, \num{0.99}) \\
    \cellcolor{white} & (2000, 100) & 0.923 & \bfseries 0.011 & \bfseries (\num{0.61}, \num{0.93}) & \bfseries 0.028 & \bfseries (\num{0.72}, \num{0.97}) \\
    \cellcolor{white} & (2000, 1000) & 0.929 & \bfseries 0.009 & \bfseries (\num{0.67}, \num{0.95}) & \bfseries 0.028 & \bfseries (\num{0.72}, \num{0.97}) \\
    \multirow{-5}{*}{\cellcolor{white} STNN} & (4000, 2000, 1000, 1000) & 0.924 & \bfseries 0.010 & \bfseries (\num{0.67}, \num{0.95}) & \bfseries 0.028 & \bfseries (\num{0.67}, \num{0.95}) \\
    \midrule
    \cellcolor{white} & (1000) & 0.931 & \bfseries 0.021 & \bfseries (\num{0.85}, \num{1.00}) & \bfseries 0.040 & \bfseries (\num{0.85}, \num{1.00}) \\
    \cellcolor{white} & (4000) & 0.932 & \bfseries 0.021 & \bfseries (\num{0.85}, \num{1.00}) & \bfseries 0.043 & \bfseries (\num{0.85}, \num{1.00}) \\
    \cellcolor{white} & (2000, 100) & 0.931 & \bfseries 0.019 & \bfseries (\num{0.72}, \num{0.97}) & \bfseries 0.035 & \bfseries (\num{0.85}, \num{1.00}) \\
    \cellcolor{white} & (2000, 1000) & 0.940 & \bfseries 0.028 & \bfseries (\num{0.85}, \num{1.00}) & \bfseries 0.052 & \bfseries (\num{0.85}, \num{1.00}) \\
    \multirow{-5}{*}{\cellcolor{white} U-MTNN} & (4000, 2000, 1000, 1000) & 0.939 & \bfseries 0.025 & \bfseries (\num{0.85}, \num{1.00}) & \bfseries 0.045 & \bfseries (\num{0.85}, \num{1.00}) \\
    \midrule
    \cellcolor{white} & (1000) & 0.935 & \bfseries 0.019 & \bfseries (\num{0.85}, \num{1.00}) & \bfseries 0.041 & \bfseries (\num{0.85}, \num{1.00}) \\
    \cellcolor{white} & (4000) & 0.934 & \bfseries 0.020 & \bfseries (\num{0.85}, \num{1.00}) & \bfseries 0.044 & \bfseries (\num{0.85}, \num{1.00}) \\
    \cellcolor{white} & (2000, 100) & 0.933 & \bfseries 0.023 & \bfseries (\num{0.78}, \num{0.99}) & \bfseries 0.042 & \bfseries (\num{0.85}, \num{1.00}) \\
    \cellcolor{white} & (2000, 1000) & 0.941 & \bfseries 0.029 & \bfseries (\num{0.85}, \num{1.00}) & \bfseries 0.051 & \bfseries (\num{0.85}, \num{1.00}) \\
    \multirow{-5}{*}{\cellcolor{white} W-MTNN} & (4000, 2000, 1000, 1000) & 0.939 & \bfseries 0.025 & \bfseries (\num{0.85}, \num{1.00}) & \bfseries 0.048 & \bfseries (\num{0.85}, \num{1.00}) \\
    \bottomrule
    \end{tabular}
\end{table*}

\begin{table*}[htbp]
    \caption{Comparisons between neural network models using random cross-validation
    with target step evaluation. Differences between STNN, U-MTNN, and W-MTNN
    models with the same core architecture are reported as median $\Delta$AUC
    values and sign test 95\% confidence intervals. Bold values indicate
    confidence intervals that do not include 0.5.}
    \label{appendix:table:random_mtnn_vs_stnn_alternative}
    \centering
    \rowcolors{2}{lightgray}{}
    \sisetup{detect-all=true}
    \begin{tabular}{ l l S c S c }
    \toprule
     & & \multicolumn{2}{c}{MTNN - STNN} &
         \multicolumn{2}{c}{W-MTNN - U-MTNN} \\
    \cmidrule(lr){3-4} \cmidrule(lr){5-6}
     & Model &
    {\makecell{Median \\ $\Delta$AUC}} & \makecell{Sign Test \\ 95\% CI} &
    {\makecell{Median \\ $\Delta$AUC}} & \makecell{Sign Test \\ 95\% CI} \\
    \midrule
    \cellcolor{white} & (1000) & \bfseries 0.010 & \bfseries (\num{0.61}, \num{0.93}) &  &  \\
    \cellcolor{white} & (4000) & \bfseries 0.009 & \bfseries (\num{0.72}, \num{0.97}) &  &  \\
    \cellcolor{white} & (2000, 100) & \bfseries 0.013 & \bfseries (\num{0.52}, \num{0.87}) &  &  \\
    \cellcolor{white} & (2000, 1000) & \bfseries 0.017 & \bfseries (\num{0.85}, \num{1.00}) &  &  \\
    \multirow{-5}{*}{\cellcolor{white} U-MTNN} & (4000, 2000, 1000, 1000) & \bfseries 0.015 & \bfseries (\num{0.78}, \num{0.99}) &  &  \\
    \midrule
    \cellcolor{white} & (1000) & \bfseries 0.010 & \bfseries (\num{0.57}, \num{0.90}) & 0.001 & (\num{0.47}, \num{0.84}) \\
    \cellcolor{white} & (4000) & \bfseries 0.008 & \bfseries (\num{0.72}, \num{0.97}) & 0.000 & (\num{0.39}, \num{0.77}) \\
    \cellcolor{white} & (2000, 100) & \bfseries 0.013 & \bfseries (\num{0.52}, \num{0.87}) & 0.006 & (\num{0.47}, \num{0.84}) \\
    \cellcolor{white} & (2000, 1000) & \bfseries 0.016 & \bfseries (\num{0.85}, \num{1.00}) & -0.000 & (\num{0.27}, \num{0.65}) \\
    \multirow{-5}{*}{\cellcolor{white} W-MTNN} & (4000, 2000, 1000, 1000) & \bfseries 0.015 & \bfseries (\num{0.67}, \num{0.95}) & 0.001 & (\num{0.35}, \num{0.73}) \\
    \bottomrule
    \end{tabular}
\end{table*}

\begin{table*}[htb]
    \caption{Pairwise comparisons between neural network model architectures
    using random cross-validation with target step evaluation. For each pair of
    models within a model class (e.g.~STNN), we report the median
    $\Delta$AUC and sign test 95\% confidence interval. Bold values indicate
    confidence intervals that do not include 0.5.}
    \label{appendix:table:random_architecture_comparison_alternative}
    \centering
    \rowcolors{1}{}{lightgray}
    \sisetup{detect-weight=true,detect-inline-weight=math}
    \begin{tabular}{ l l l S c }
    \toprule
     & & & \multicolumn{2}{c}{Model B - Model A} \\
    \cmidrule(lr){4-5}
     & Model A & Model B &
    {\makecell{Median \\ $\Delta$AUC}} &
    \makecell{Sign Test \\ 95\% CI} \\
    \midrule
    \cellcolor{white} & (1000) & (4000) & \bfseries 0.002 & \bfseries (\num{0.67}, \num{0.95}) \\
    \cellcolor{white} & (1000) & (2000, 100) & 0.000 & (\num{0.39}, \num{0.77}) \\
    \cellcolor{white} & (1000) & (2000, 1000) & -0.000 & (\num{0.27}, \num{0.65}) \\
    \cellcolor{white} & (1000) & (4000, 2000, 1000, 1000) & 0.001 & (\num{0.35}, \num{0.73}) \\
    \cellcolor{white} & (4000) & (2000, 100) & \bfseries \color{red} -0.001 & \bfseries \color{red} (\num{0.10}, \num{0.43}) \\
    \cellcolor{white} & (4000) & (2000, 1000) & \bfseries \color{red} -0.001 & \bfseries \color{red} (\num{0.10}, \num{0.43}) \\
    \cellcolor{white} & (4000) & (4000, 2000, 1000, 1000) & \bfseries \color{red} -0.001 & \bfseries \color{red} (\num{0.10}, \num{0.43}) \\
    \cellcolor{white} & (2000, 100) & (2000, 1000) & 0.000 & (\num{0.39}, \num{0.77}) \\
    \cellcolor{white} & (2000, 100) & (4000, 2000, 1000, 1000) & 0.000 & (\num{0.35}, \num{0.73}) \\
    \multirow{-10}{*}{\cellcolor{white} STNN} & (2000, 1000) & (4000, 2000, 1000, 1000) & -0.000 & (\num{0.23}, \num{0.61}) \\
    \midrule
    \cellcolor{white} & (1000) & (4000) & 0.001 & (\num{0.47}, \num{0.84}) \\
    \cellcolor{white} & (1000) & (2000, 100) & -0.003 & (\num{0.20}, \num{0.57}) \\
    \cellcolor{white} & (1000) & (2000, 1000) & \bfseries 0.009 & \bfseries (\num{0.78}, \num{0.99}) \\
    \cellcolor{white} & (1000) & (4000, 2000, 1000, 1000) & \bfseries 0.005 & \bfseries (\num{0.57}, \num{0.90}) \\
    \cellcolor{white} & (4000) & (2000, 100) & -0.002 & (\num{0.27}, \num{0.65}) \\
    \cellcolor{white} & (4000) & (2000, 1000) & \bfseries 0.008 & \bfseries (\num{0.85}, \num{1.00}) \\
    \cellcolor{white} & (4000) & (4000, 2000, 1000, 1000) & \bfseries 0.005 & \bfseries (\num{0.57}, \num{0.90}) \\
    \cellcolor{white} & (2000, 100) & (2000, 1000) & \bfseries 0.010 & \bfseries (\num{0.78}, \num{0.99}) \\
    \cellcolor{white} & (2000, 100) & (4000, 2000, 1000, 1000) & \bfseries 0.005 & \bfseries (\num{0.72}, \num{0.97}) \\
    \multirow{-10}{*}{\cellcolor{white} U-MTNN} & (2000, 1000) & (4000, 2000, 1000, 1000) & -0.002 & (\num{0.20}, \num{0.57}) \\
    \midrule
    \cellcolor{white} & (1000) & (4000) & -0.000 & (\num{0.27}, \num{0.65}) \\
    \cellcolor{white} & (1000) & (2000, 100) & 0.001 & (\num{0.35}, \num{0.73}) \\
    \cellcolor{white} & (1000) & (2000, 1000) & \bfseries 0.009 & \bfseries (\num{0.72}, \num{0.97}) \\
    \cellcolor{white} & (1000) & (4000, 2000, 1000, 1000) & \bfseries 0.007 & \bfseries (\num{0.67}, \num{0.95}) \\
    \cellcolor{white} & (4000) & (2000, 100) & 0.000 & (\num{0.35}, \num{0.73}) \\
    \cellcolor{white} & (4000) & (2000, 1000) & \bfseries 0.007 & \bfseries (\num{0.72}, \num{0.97}) \\
    \cellcolor{white} & (4000) & (4000, 2000, 1000, 1000) & \bfseries 0.004 & \bfseries (\num{0.57}, \num{0.90}) \\
    \cellcolor{white} & (2000, 100) & (2000, 1000) & \bfseries 0.006 & \bfseries (\num{0.67}, \num{0.95}) \\
    \cellcolor{white} & (2000, 100) & (4000, 2000, 1000, 1000) & \bfseries 0.004 & \bfseries (\num{0.85}, \num{1.00}) \\
    \multirow{-10}{*}{\cellcolor{white} W-MTNN} & (2000, 1000) & (4000, 2000, 1000, 1000) & -0.002 & (\num{0.23}, \num{0.61}) \\
    \bottomrule
    \end{tabular}
\end{table*}

\begin{figure}[tb]
  \centering
  \includegraphics[width=0.5\linewidth]{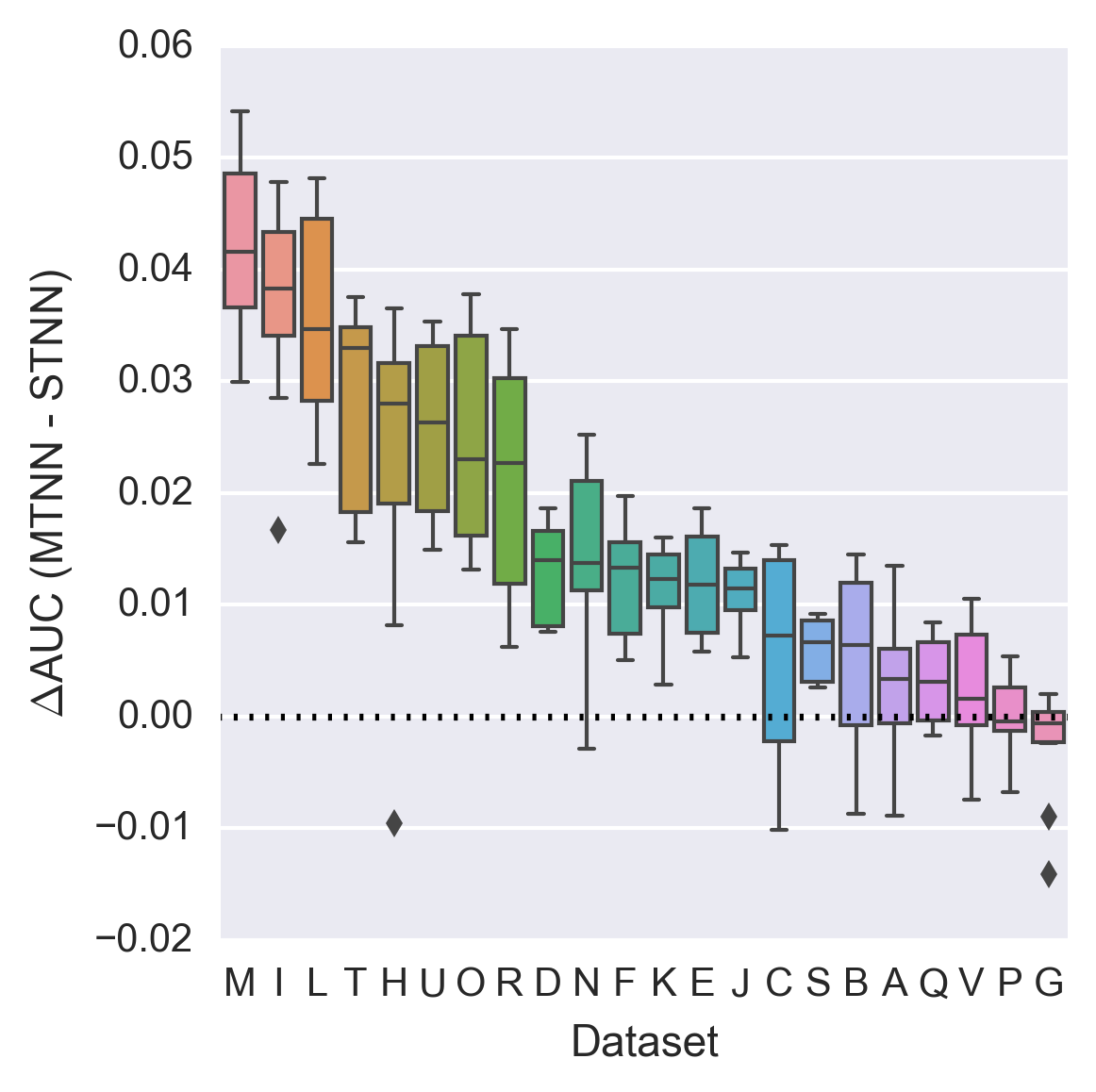}
  \caption{Box plots showing $\Delta$AUC values between MTNN
  and STNN models with the same core architecture for models trained
  using random cross-validation with target step evaluation. Each box plot
  summarizes 10 $\Delta$AUC values, one for each combination of model
  architecture (e.g.~(2000,~1000)) and task weighting strategy (U-MTNN or
  W-MTNN).}
  \label{fig:multitask_effect_box_random_best}
\end{figure}

\putbib
\end{bibunit}

\end{document}